\documentclass[lettersize,journal]{IEEEtran}
\usepackage{amsmath,amsfonts}
\usepackage{algorithm}
\usepackage{array}
\usepackage[caption=false,font=normalsize,labelfont=sf,textfont=sf]{subfig}
\usepackage{textcomp}
\usepackage{stfloats}
\usepackage{url}
\usepackage{verbatim}
\usepackage{graphicx}
\usepackage{cite}
\usepackage{microtype}
\usepackage{graphicx}
\usepackage{booktabs} 

\usepackage{hyperref}
\usepackage{algpseudocode}
\usepackage{algorithm}
\usepackage{dashrule}
\usepackage{arydshln}
\usepackage{spverbatim}
\usepackage{tikz}
\usepackage{color}
\usepackage{booktabs}
\usepackage{multirow}
\usepackage{amssymb}
\usepackage{amsmath} 
\usepackage{mathabx}
\usepackage{graphicx}
\usepackage{color,xcolor}
\usepackage{mathrsfs}
\usepackage{fancyhdr}
\usepackage{eucal}
\usepackage{stmaryrd}
\usepackage{dashrule}
\usepackage{tikz}
\usepackage{xcolor}

\begin{document}

\title{CausalGPT: Illuminating Faithfulness and Causality
for Knowledge Reasoning with Foundation Models}

\author{Ziyi Tang, Ruilin Wang, Weixing Chen, Yongsen Zheng, Zechuan Chen, Yang Liu, Keze Wang, Tianshui Chen, Liang Lin,~\IEEEmembership{Fellow,~IEEE,}
\thanks{Ziyi Tang, Ruilin Wang, Weixing Chen, Yongsen Zheng, Zechuan Chen, Yang Liu, Keze Wang, and Liang Lin are with the Sun Yat-sen University, Guangzhou, Guangdong 510275, China (email: tangzy27@mail2.sysu.edu.cn; linliang@ieee.org).}
}

\markboth{ IEEE TRANSACTIONS ON PATTERN ANALYSIS AND MACHINE INTELLIGENCE,~Vol.~14, No.~8, August~2021}%
{Shell \MakeLowercase{\textit{et al.}}: A Sample Article Using IEEEtran.cls for IEEE Journals}


\maketitle
\begin{abstract}
Despite the progress of foundation models, knowledge-based reasoning remains a persistent challenge due to their limited capacity for knowledge recall and inference. Existing methods primarily focus on encouraging these models to plan and solve problems or extensively sample reasoning chains independently. However, these methods often overlook conceptual errors and inferential fallacies, inevitably leading to a series of notorious issues such as misleading conclusions, cognitive biases, and reduced decision quality. {\color{black} While explicit modeling of causality is argued to hold promise in addressing these issues, contemporary research efforts have thus far fallen short in achieving causality-based foundation models. Drawing inspiration from the orchestration of diverse specialized agents collaborating to tackle intricate tasks, we propose a framework named Causal-Consistency Chain-of-Thought (CaCo-CoT) that harnesses multi-agent collaboration to bolster the faithfulness and causality of foundation models}, involving a set of \textit{reasoners} and \textit{evaluators}. 
These agents collaboratively work within a reasoning-and-consensus paradigm to improve faithfulness. The \textit{reasoners} are tasked with generating reasoning chains for knowledge-intensive problems by mimicking human causal reasoning. {\color{black}Meanwhile, the \textit{evaluator} scrutinizes the causal consistency of a \textit{reasoner}'s reasoning chain from a non-causal and a counterfactual perspective}. Our framework demonstrates significant superiority over state-of-the-art methods through extensive and comprehensive evaluations across text-based and multi-modal knowledge reasoning tasks (e.g., science question answering and commonsense reasoning).
\end{abstract}


\section{Introduction}
\IEEEPARstart{R}{ecent} advancements in foundation models usher in a new era. By assimilating vast amounts of global knowledge and incorporating human feedback, these models - particularly language models and multi-modal models - are evolving into colossal knowledge warehouses~\cite{wang-etal-2024-knowledge-mechanisms, roberts-etal-2020-much}. Reasoning serves as the key to unlocking the true value of this accumulated knowledge, enabling its effective utilization in practical applications. Starting from known knowledge, reasoning is a complicated cognitive process, employing various modes of thinking to arrive at faithful conclusions~\cite{huang-chang-2023-towards}. The core issue with reasoning in foundation models lies in their limited capability to guarantee the faithfulness of conclusions~\cite{zhao2023survey}, as evidenced by their struggles with difficulties in identifying nonfactual knowledge and inferential errors~\cite{huangetal2023zero}. Thus, the capability of reasoning, a fundamental aspect of intelligence, continues to pose a longstanding challenge. 

{\color{black} Many research efforts have been dedicated to eliciting autonomous reasoning in foundation models by mimicking human-like step-wise reasoning processes from multiple given demonstrations~\cite{wei2022chain, yan2024understanding, Liu_Qin_Ye_Mou_He_Wang_2024} or creating an extensive ensemble of reasoning chains~\cite{wang2022self, fu2022complexity, yoran-etal-2023-answering, yao2023tree}. Besides, some other approaches~\cite{wang2023plan, press2022measuring} allow these models to autonomously plan and solve sub-problems or dynamically investigate with more complex reasoning structures~\cite{yao2023beyond, besta2023got, yao2023tree}. More recently, multi-agent frameworks have been engineered to empower foundation models to autonomously tackle the complex challenges of the real world~\cite{Shah_Conrad_Williams_2009, du2023improving, li2023camel, talebirad2023multiagent, wang-wong-2021-collaborative, Du_2023, weize2024agentverse}, e.g., software development.}

\begin{figure}[!t]
%
\center
\includegraphics[width=0.488\textwidth]{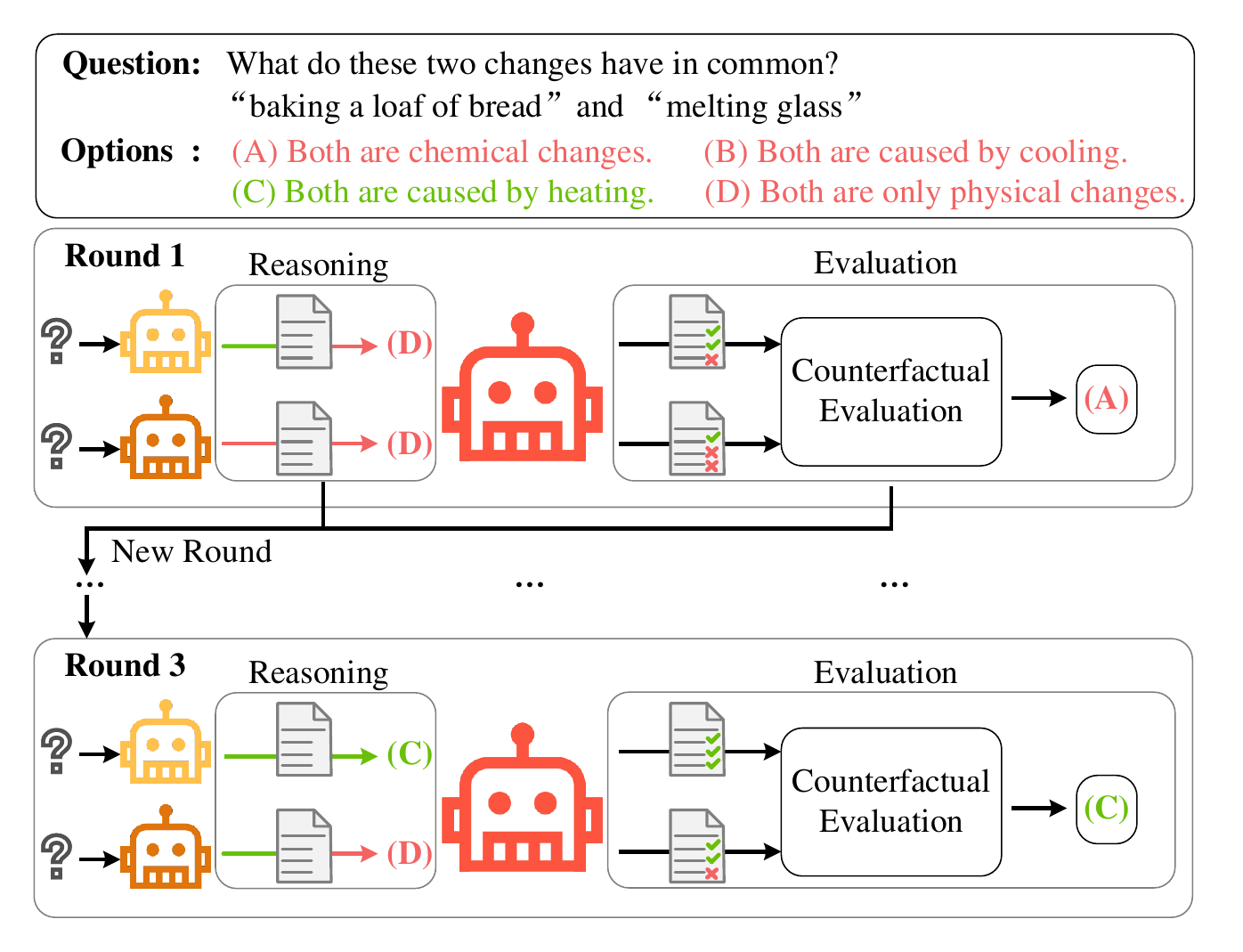}
\caption{The overview of our CaCo-CoT. Given a question from ScienceQA, \textit{reasoners} fail to capture the golden answer (C) in the first cooperation round. For this issue, multi-round cooperation between multiple \textit{reasoners} and \textit{evaluators} is adopted to yield a faithful answer. 
} 
\label{fig:intro}
\end{figure}

Despite their effectiveness, these methods still suffer from two major issues. \emph{1) Reasoning Mechanism}. They lack explicit mechanisms to effectively address the core issues of ensuring knowledge factualness and inference errors that lead to cascading errors in reasoning chains~\cite{chen2023hallu,huangetal2023zero,li-du-2023-leveraging,li2023halueval}. Consequently, the inference outcomes generated by these systems inevitably result in a range of notorious problems such as misleading conclusions, cognitive biases, and reduced decision quality. These constraints have notably reduced the precision of foundation model-based autonomous reasoning systems and imposed limitations on their practical utility across diverse domains. \emph{2) Collaborative Strategy}. Despite advancements, a plethora of challenges persists regarding agent cooperation, including the emergence of cascading hallucinations~\cite{hong2024metagpt} and the risk of infinite loops~\cite{li2023camel}. These issues not only reduce token efficiency but also undermine the efficacy of knowledge reasoning processes. Addressing these challenges requires innovative approaches to foster seamless collaboration among agents, ensuring reliable information exchange and mitigating the risk of computational inefficiencies that can impede successful reasoning outcomes.

As shown in Fig. \ref{fig:intro}, the focus of this paper is devising a multi-agent system that effectively handles potential factual and inferential errors in knowledge reasoning. CaCo-CoT employs a hierarchical arrangement of multiple agents, specifically, a set of faithful \textit{reasoner} agents and causal \textit{evaluator} agents, simplified as \textit{reasoners} and \textit{evaluators}, respectively. 
To deduce an answer, \textit{reasoners} are devoted to executing a sequence of sub-steps, modeled after the human reasoning process. 
Subsequently, the \textit{evaluator} is responsible for assessing the causal consistency of each top-ranked reasoning chain, ensuring that the selected chains adhere to logical thinking and causal principles. 
Once consensus is achieved within the agent system, the agreed-upon answer is adopted as the final decision.
If not, the cooperation process persists recursively, until a consensus is reached. 
Specifically, the task for the \textit{reasoners} and the \textit{evaluator} is multifaceted. 
The \textit{reasoners} engage in reasoning within a structure that exhibits sequential dependence, comprising sub-steps of \textit{concept explanation}, \textit{sub-question decomposition}, and \textit{sub-question answering}. 
To verify the faithfulness of a reasoning chain, the \textit{evaluator}, first scrutinizes the statements therein one by one referring to the whole reasoning structure, which follows a non-causal sequence contrary to the reasoning chain's causal inference direction. 
Then, the \textit{evaluator} moves on to counterfactual evaluation where it considers a counterfactual premise and distinguishes any contradictory evidence. 

Overall, our \textbf{main} contributions in this paper are three-fold:
\begin{itemize}
\setlength{\itemsep}{0pt}
\setlength{\parsep}{0pt}
\setlength{\parskip}{0pt}
  \item {\color{black} To the best of our knowledge, this is the first work to illuminate faithfulness and causal consistency in foundation models via a multi-agent collaborative paradigm for knowledge-based reasoning. This can remarkably improve the reasoning faithfulness of foundation models.}
  
  \item {\color{black}We proposed a novel framework, CaCo-CoT, which implements two types of agents, i.e., faithful \textit{reasoner} and causal \textit{evaluator} (or \textit{reasoner} and \textit{evaluator}). The former aims to faithfully derive reasoning chains for knowledge reasoning problems with the latter to evaluate these reasoning chains based on causal consistency.}
  
  \item Extensive experiments demonstrate that CaCo-CoT achieves state-of-the-art performance on text-based and multi-modal knowledge reasoning benchmarks, revealing the effectiveness of illuminating faithfulness and causality for knowledge-based reasoning in foundation models.
  
\end{itemize}

\section{Related Work}
\subsection{Reasoning Ability of Foundation Models}

As foundation models scale up in terms of data and parameters, they exhibit emergent abilities - capabilities that are absent in smaller models but naturally arise in larger ones. These emergent phenomena include arithmetic reasoning~\cite{chung2022scaling}, chain-of-thought (CoT) reasoning~\cite{wei2022chain}, and multi-modal reasoning~\cite{proreason, openai2023gpt4}, enabling foundation models to tackle increasingly complex tasks~\cite{wei2022emergent}.
However, these exceptional foundation models still have limitations, such as producing hallucinations with factual errors or potential risk responses in certain specific contexts~\cite{openai2023gpt4, bang2023multitask}. 
Essentially, foundation models appear to unconsciously utilize knowledge to solve tasks, yet cannot still control the accuracy of knowledge usage~\cite{zhao2023survey}.
Firstly, subsequent studies focus on enhancing CoT's potential to improve reasoning accuracy \cite{fu2022complexity}, including diverse reasoning paths sampling \cite{wang2022self}, sub-problems divided for sequential solving \cite{zhou2022least}, introducing self-asking and self-answering \cite{press2022measuring}, and employing multi-agent debates \cite{du2023improving}. 
However, they do not focus on identifying errors in the reasoning paths.
Secondly, several works tried to strengthen, update, and discriminate foundation models' responses by introducing external knowledge. 
This external knowledge can be injected into foundation models by providing them with relevant dense vectors or documents retrieved from the knowledge base~\cite{lewis2020retrieval,borgeaud2022improving}.
%
Similarly, it can also be formatted into logical forms, combined with responses in natural language to obtain the final answer~\cite{yu2022decaf}. 
The performance of this method is significantly constrained by the capabilities of the information retrieval system and the quality of the knowledge base.
Furthermore, in the multi-modal reasoning domain, CompCoT~\cite{MitraCCoT} leverages scene graph representations to enhance the compositional reasoning capabilities of multi-modal foundation models through a zero-shot chain-of-thought prompting approach. 
Zhang et al. \cite{zhang2024multimodalchainofthoughtreasoninglanguage} propose a two-stage framework that separates rationale generation and answer inference, demonstrating that explicit modality fusion in the reasoning process can mitigate hallucination even with relatively small model sizes.


\subsection{Causality in Foundation Models}

The causal abilities of foundation models are a highly debated issue~\cite{zevcevic2023causal, jin2024can}, but it is undeniable that foundation models have made significant improvements on multiple causal benchmarks~\cite{kıcıman2023causal}. Additionally, traditional causal methods heavily rely on potential causal mechanisms and prior domain knowledge within a system that cannot be explicitly demonstrated~\cite{stolfo-etal-2023-causal}. Furthermore, causal knowledge in traditional causal methods is difficult to transfer to other domains, limiting the practicality of causal methods. In contrast,  foundation models utilize background knowledge about the real world as prior information, enabling them to comprehend and formalize diverse causal scenarios and use this knowledge to make accurate reasoning about unseen events~\cite{chen2023mitigating}.
Specifically, the emergence capability of foundation models can be leveraged to perform causal discovery using the internal knowledge of the model. In \cite{long2022can}, foundation models are used to generate causal structures for a limited set of variables, relying on the connectors between variables given in the text. 
Therefore, \cite{ban2023query} further integrated the generated causal knowledge as input, enabling foundation models to discover more faithful and novel causal relationships from data, and integrated foundation models as a causal knowledge querying tool into classical models, bringing significant performance improvement.

\section{Causal-consistency Chain-of-Thought}

The primary objective of this paper is to address the constraints inherent in modern foundation models through the integration of causality principles, with the ultimate goal of achieving CausalGPT, a groundbreaking approach that harnesses causal reasoning for superior prediction and decision-making.
In particular, our focus narrows to tackling knowledge-based reasoning (KR) problems using foundation models. These problems are usually articulated in natural language or multi-modal data (such as text and images). 
%
%
Addressing these problems necessitates background knowledge, often not provided explicitly, serving as a premise for reasoning. 
Therefore, foundation models have to retrieve knowledge from their internal knowledge space and conduct knowledge inferring~\cite{wang-etal-2024-knowledge-mechanisms, roberts-etal-2020-much, jin2024exploringconceptdepthlarge}, i.e., to employ acquired knowledge to derive unknown knowledge. 

\begin{figure*}[!t]
\center
\includegraphics[width=\textwidth]{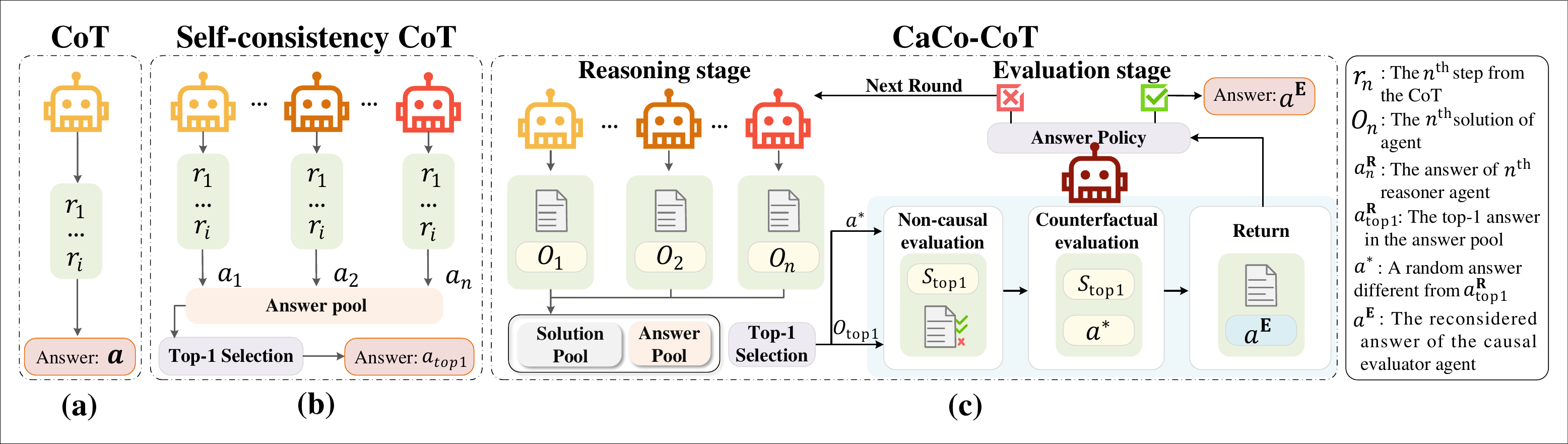}

\caption{Comparison between existing approaches and causal-consistency chain-of-thought (CaCo-CoT). CaCo-CoT introduces a collaborative mechanism where faithful reasoner agents (upper left) and causal evaluator agents (right) cooperate to produce a causally consistent reasoning chain, aiming to minimize factual and inferential errors. } 
\label{fig:method}
\vspace{-10pt}
\end{figure*}

%


The core idea of the causal consistency chain-of-thought (CaCo-CoT) framework is to resolve KR problems through a multi-agent decision-making process where each agent plays the role of either a faithful reasoner (\textit{reasoner} for short) or a causal evaluator (\textit{evaluator} for short). 
Each agent is implemented by prompting a foundation model with in-context learning capability to follow our agent's instruction. 
In this section, we first delineate how a \textit{reasoner} strives to tackle a KR problem within a structured reasoning procedure. 
The next subsection introduces the \textit{evaluator}, which assesses the reasoning process in the reasoning chain $O$ collected from a \textit{reasoner} $\textbf{R}$. 
This evaluation is accomplished by scrutinizing the causal consistency, including non-causal evaluation and counterfactual evaluation. 
In the end, the proposed multi-agent cooperation framework, CaCo-CoT, is delineated. 
%

\subsection{Faithful Reasoner} \label{sec:reasoner}

\begin{figure*}[!t]
\center
\includegraphics[width=\textwidth]{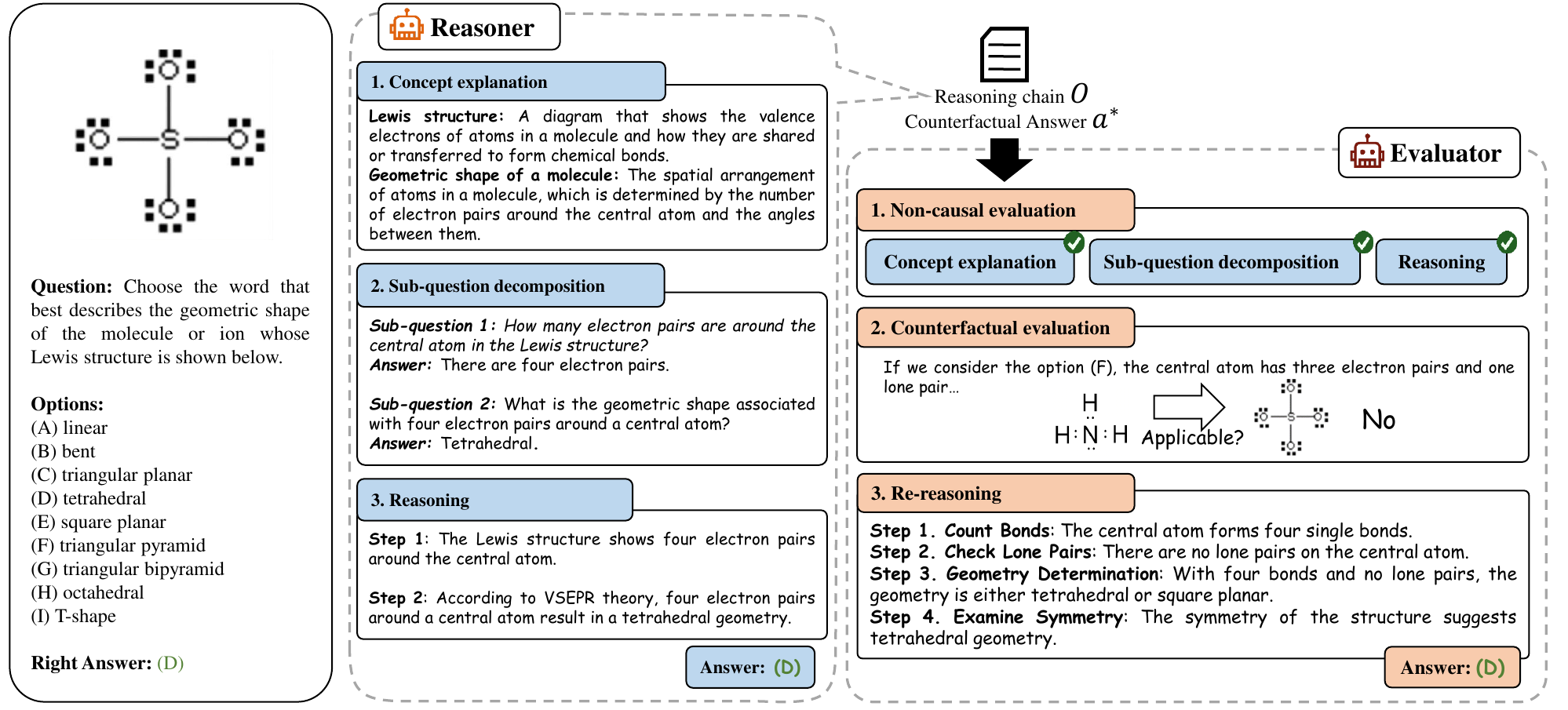}

\caption{Demonstration of how a \textit{faithful reasoner} and a \textit{causal evaluator} analyze a molecular geometry question. } 
\label{fig:3}
\vspace{-10pt}
\end{figure*}

Prior works demonstrate that foundation models benefit from mimicking human minds to solve complex problems~\cite{wei2022chain, yang2024buffer, zhou2022least}. 
One of the pathways that human minds solve knowledge reasoning tasks can boil down to an underlying causal procedure, which is usually neglected but more universal: 1) understanding the question; 2) retrieving the pertinent knowledge; 3) planning how to solve it step by step; 4) following the plan to reach the answer. 
Here, each sub-task partially hinges on how previous sub-tasks are finished, but not vice versa, i.e., this procedure is directed. 
This is because the results of previous sub-tasks are the prerequisite to accomplishing the next sub-task. 
%
%
The process by which humans fulfill each sub-task is faithful since they get down to each sub-task only when they have obtained sufficient premises. 
In contrast, foundation models are usually misled by unstructured reasoning~\cite{zhou2024selfdiscoverlargelanguagemodels, yang2024buffer, qi2024mutualreasoningmakessmaller}. 
This process of human cognition can be conceptualized as a causal procedure, where each step, represented as a node, maintains a directed causal relationship with other nodes. 

This causal procedure is analogous to auto-regression~\cite{radford2019language, radford2018improving}, the foundation of contemporary foundation models. 
Motivated by this analogy, it is reasonable to anticipate that a \textit{reasoner} programmed to align with the intrinsic patterns of human causal reasoning may yield faithful outcomes.
Let $i$ denote the total sub-step number to solve a multi-modal question $Q$ that may contain multi-modal tokens. To clarify how a foundation model-based \textit{reasoner} agent functions, we formulate the auto-regressive generation process of a sub-step $r_{i^\prime} \in \{1, \ldots, i\}$ for $Q$ as follows: 

\begin{equation}
\begin{aligned}
r_{i^\prime} = f_{\theta, P_\textbf{R}}(Q, (r_1, \ldots, r_{i^\prime-1})) \\
\end{aligned}
\label{eq:1}
\end{equation}

\noindent where $f_{\theta, P_\textbf{R}}$ represents a foundation model $f$, typically a Transformer decoder~\cite{c:22, radford2019language}, with pre-trained parameters $\theta$ and the prompt $P_\textbf{R}$. Let $(.)$ denote an ordered set and $\{.\}$ an unordered set. When $f_{\theta, P_\textbf{R}}$ is invoked, the model performs chat completion by employing auto-regressive decoding, whereby it sequentially predicts the subsequent tokens based on the preceding context, iteratively generating text until an \texttt{end-of-sequence} token is encountered. 
$P_\textbf{R}$ and $Q$ both consist of a sequence of tokens. 
%
%

As described in Eq. \ref{eq:1}, the generation of the $i^{\prime \, (\text{th})}$ sub-step $r_{i^{\prime}}$ only refer to its parent sub-steps $\{r_1, \ldots, r_{i^\prime-1}\}$. 
Built on auto-regressive generation, \textit{reasoner} agents can be implemented to mimic human causality, a directed causal generation process, by sequentially executing an instruction in the prompt $P_\textbf{R}$ for KR problems. 
We define the faithful reasoner by $\textbf{R}:= f_{\theta, P_{\textbf{R}}}$, where `$:=$' means ``is defined as''. It can be viewed as a foundation model endowed with instructions for knowledge reasoning. 

%
%


%

These instructions are given by the prompt $P_\textbf{R}$, a sequence of tokens for agent $\textbf{R}$ to implement some abilities, which can be invoked by their corresponding functions, e.g., $\mathrm{Expl}(.)$. 
%
We now elaborate the specific reasoning process of the \textit{reasoners} with the sub-step number $i = 3$. 

Initially, the first sub-step is to clarify concepts pertinent to $Q$, such as technical terms, principles, etc. This sub-step demands a foundation model to recall necessary conceptual knowledge directly from its pre-trained knowledge space, without using the retrieval augmented generation (RAG) technique. This sub-step can be described as:

\begin{equation}
\begin{aligned}
r_1 := \{(c_1, x_1) \ldots, (c_m, x_m)\} = \mathrm{Expl}(\textbf{R}, Q), \\
\quad \textbf{s.t.} \quad Q \models \{c_1, \ldots, c_m\} 
\end{aligned}
\label{eq:2}
\end{equation}



\noindent where $\mathrm{Expl}(.)$ represents a function of the \textit{reasoner} that identifies and explains the possible concepts respectively through prompting its back-end foundation model. The symbol $\models$ represents the logical symbol ``\texttt{MODEL}'', making $\{c_1, \ldots, c_m\}$ an unordered set of pertinent concepts modeled by the question $Q$. These concepts are explained one-by-one by the \textit{reasoner} \textbf{R}.  
This sub-step motivates the \textit{reasoner} to have an unbiased understanding of the prerequisite knowledge used in the question and thereby minimizes the likelihood of misinterpretation.

Secondly, the \textit{reasoner} sub-question decomposition, which breaks down the main question into a sequence of $t$ atomic sub-questions $(q_1, \ldots, q_t)$, leveraging the retrieved knowledge in $r_1$: 

\vspace{-10pt}
\begin{equation}
\begin{aligned}
r_2 := (q_1, \ldots, q_t) = \mathrm{Decomp}(\textbf{R}, Q, r_1) 
\end{aligned}
\label{eq:3}
\end{equation}

%
\noindent where $\mathrm{Decomp}(.)$ is to activate the decomposition ability of the \textit{reasoner} $\textbf{R}$ with succinct prompts, e.g., ``\texttt{Decompose the question into several subquestions connected logically to arrive at the final answer}''. 
These sub-questions $(q_1, \ldots, q_t)$ are also generated in an auto-regressive manner for better logical coherence, to approach the final answer step-by-step. 
Having been proven effective~\cite{wei2022chain, zhou2022least}, problem decomposition ensures that each atomic sub-question is sufficiently specific and manageable for foundation model-based agents, achieving high success rates. 
%
%

Next, the agent needs to solve each sub-question one by one to derive intermediate evidence for the final answer, considering previous sub-answers as evidence, which can be formulated as: 


\vspace{-10pt}
\begin{equation}
\begin{aligned}
r_3 &:= \{e_1, \ldots, e_t\}, \\
\text{s.t.} \ e_{t^{\prime}} = \mathrm{Solve}\big(&\textbf{R}, q_t, (r_1, r_2, \{e_1, \ldots, e_{{t^{\prime}}-1}\})\big). 
\end{aligned}
\label{eq:4}
\end{equation}

Finally, previous results $\{r_1, r_2, r_3\}$ are integrated to deduce a comprehensive answer $a^\textbf{R}$ to the original question $Q$, i.e., 

\vspace{-5pt}
\begin{equation}
\begin{aligned}
a^\textbf{R} = \mathrm{Solve}\big(\textbf{R}, Q, (r_1, r_2, r_3)\big).
\end{aligned}
\label{eq:5}
\end{equation}

\noindent And all sub-step results are concatenated to obtain the reasoning chain $O:= \{r_1, r_2, r_3, a^\textbf{R}\}$.



%
%

In summary, in Eq. \ref{eq:1}-\ref{eq:5}, the \textit{reasoner} agent $\textbf{R}$ sequentially generates intermediate results for a question $Q$, including $r_1$ for concept explanation, $r_2$ for sub-question decomposition, and $r_3$ for sub-question answering, finally reaching the answer $a^\textbf{R}$ and composing reasoning chain $O$. This sequential deduction process, wherein each $r_{i^\prime} \in \{r_1, \ldots, r_i\}$  is inferred based on its antecedent states $(r_1, \ldots, r_{i^\prime-1})$, exemplifies how humans utilize causality for addressing knowledge-based questions. 
%

\subsection{Causal Evaluator}

Foundation models are capable of recognizing both factual or inferential mistakes in text~\cite{li2023halueval, madaan2023selfrefine, weng2023large}. Yet, previous works primarily focus on identifying the presence of these errors, without pinpointing the specific location or type of the errors identified, such as misguided reasoning directions and false reasoning evidence. This lack of detail renders these found errors less faithful. 
To ensure faithful reasoning, we introduce a causal evaluator agent or simply \textit{evaluator}, aimed to alleviate both factual and inferential errors within potential reasoning chains gathered from \textit{reasoners}. 
Analogously, the evaluator $\textbf{E}$, a foundation model characterized by its model weights $\theta$ and an instructive prompt $P_\textbf{E}$, can be represented by the function $f_{\theta, P_\textbf{E}}$, defined as $\textbf{E}$, that is, $\textbf{E} := f_{\theta, P_\textbf{E}}$.

The \textit{evaluator} examines causal consistency within a reasoning chain $O$ as working from a non-causal perspective and a counterfactual perspective. 
Here, causal consistency describes whether an answer $a^\textbf{R}$, derived from a causal direction as the \textit{reasoner} $\textbf{R}$ working in, is consistent with the answer $a^\textbf{E}$ derived by the \textit{evaluator} who scrutinizes from a non-causal and counterfactual perspective. 

We now delineate how the \textit{evaluator} examines a reasoning chain $O$ from a non-causal perspective and counterfactual perspective. 
Initially, to evaluate the factualness of knowledge and inferential soundness, the \textit{evaluator} agent conducts non-causal evaluation, that is to examine whether $r_1, r_2, r_3$ have any factual and inferential error while considering all other intermediate sub-steps. 
In this direction, the agent can detect potential factual errors in each sub-step and seek inference errors that may be neglected during the auto-regressive generation process in Eq. \ref{eq:1}. 
%
%
Non-causal evaluation can be formulated as:

\vspace{-10pt}
\begin{small}
\begin{equation}
\begin{aligned}
g_{i^\prime} = \mathrm{Eval}\big(\textbf{E}, Q, O, (g_{1}, \ldots, g_{{i^\prime-1}})\big), 
\end{aligned}
\label{eq:6}
\end{equation}
\end{small}
\vspace{-10pt}

\noindent where $g_{i^\prime} \in \{1, 0\}$ denotes $r_{i^\prime}$'s binary evaluation result, \textsc{True} or \textsc{False}. The non-causal evaluation score is calculated by $S_\text{non} = \frac{1}{i} \sum_{i^\prime=1}^{i} g_{i^\prime}$. In such a case, the evaluator needs to consider step-by-step from the beginning of the reasoning chain after the future states of this chain are already known, i.e., each sub-step $r_{i^\prime}$ is evaluated after the \textit{evaluator} has read through $(r_1, \ldots, r_i)$. The evaluator needs to consider, for example, whether a first piece of evidence has sufficient conditions to lead to the current result. This type of evaluation can be considered non-causal as it is not operated in the causal direction as the \textit{reasoner} agents do. 

It provides an opportunity to identify any emergent properties or macro-level patterns (such as dependency relationship in the rationale) that might be missed when reasoning at a micro-level (solving decomposed sub-questions one by one without considering where they may lead). 
In other words, it draws the whole inference path and inspects for any misguided directions made in reasoning. 
Overall, this step evaluates a reasoning chain $O$ from a non-causal direction different from that of its generation process, to examine its causal consistency from different causal viewpoints. 

Furthermore, to deeply detect inferential fallacies, the \textit{evaluator} is prompted to perform the counterfactual evaluation based on the evaluated counterfactual consistency, i.e., to evaluate if the original reasoning chain still holds starting from a counterfactual premise. 
%
Taking a multi-choice question as an example, in the reasoning stage, a group of \textit{reasoner} $\mathbb{R} = \{\textbf{R}_1, \ldots, \textbf{R}_n\}$ are parallelized to reason for $Q$ with a candidate answer set $c$ representing all choices associated with $Q$. 
Assume the evidence set that a group of \textit{reasoner} agents use to derive the answer pool $\{a^\textbf{R}_1, \ldots, a^\textbf{R}_n\}$, is denoted as $\textbf{e} = \{e_1, \ldots, e_{T}\}$. 
Once the answer pool contains more than one potential answer, i.e., $\exists j, k \in \{1, \ldots, n\}, a^\textbf{R}_j \neq a^\textbf{R}_k$, it can be inferred that there exist conflicts in the evidence set $\textbf{e}$ leading to different answers, i.e., $\exists j, k \in \{1, \ldots, T\}, e_j \land e_k \rightarrow \bot$ (``$\land$'': logical symbol \texttt{AND}; ``$\bot$'': logical symbol \texttt{CONTRADICTION}). 

As a result, the answers in the answer pool should be considered unfaithful. 
To examine these conflicts in $\textbf{e}$, we introduce counterfactual evaluation. 
For example, for a reasoning chain $O$ with an answer $a^\textbf{R}$, we sample another answer $a^* \neq a^\textbf{R}$ from the answer pool. 
Then, the \textit{evaluator} agent applies a counterfactual answer set $c^* = \{a^\textbf{R}, a^*\}$ as a counterfactual premise and labels each piece of evidence in the evidence set $\textbf{e}$ with the ternary values $\{b_1, \ldots, b_{T}\}$, where each element $b \in \{1, 0, -1\}$, denoting \textsc{True}, \textsc{False}, and \textsc{Irrelevant} separately. 
Concretely, the pieces of evidence marked by \textsc{Irrelevant} are excluded, 
and the remaining pieces are evaluated through self-inspection for any self-contradiction ($e_j \land e_k \rightarrow \bot$) or inference error ($\neg((Q \land c^*) \rightarrow e_j)$), ending up with a faithful evidence set $\textbf{e}^*_\textsc{True}$. 
Finally, the \textit{evaluator} agent is prompted to output a counterfactual answer $a^\textbf{E}_*$, based solely on the \textsc{True} evidence. Formally,


\begin{small}
\begin{equation}
\begin{aligned}
\textbf{e}^*_\textsc{True} = \{ e_j \mid b_j \neq & -1, \neg(e_j \land e_k \rightarrow \bot), (Q \land c^*) \rightarrow  e_j \},  \\ 
&\text{s.t.} \quad j, k \in \{1, \ldots, T\}, 
\end{aligned}
\label{eq:7}
\end{equation}
\end{small}


\begin{small}
\begin{equation}
\begin{aligned}
a^\textbf{E}_* = a^\textbf{R}_{c^*} = \mathrm{Solve}(\textbf{E}, Q, c^*, \textbf{e}^*_\textsc{True})
\end{aligned}
\label{eq:8}
\end{equation}
\end{small}

\noindent where $a^\textbf{R}_{c^*}$ means the value of answer variable $a^\textbf{R}$ in a counterfactual hypothesis that the factual candidate answer set $c$ is replaced by $c^*$. Moreover, "$\neg$" denotes logical symbols \texttt{NOT}. We leverage counterfactual consistency as a faithfulness score $S_\text{counter}$ as assigned to $a^\textbf{R}$. 
Counterfactual consistency implies whether the counterfactual answer $a^*$ is consistent with the evaluated (factual) answer $a^\textbf{R}$ when they have identical counterfactual premises~\cite{pearl2009causality}. Specifically, we relax counterfactual consistency, making it hold when 1) a factual answer $a^\textbf{R}$ can be uniquely determined; additionally, 2) it is involved in both the factual answer set premise $c$ and the counterfactual one $c^*$. In essence, $a^\textbf{R}$ should be consistent and not be unswayed by another answer if we apply counterfactual evaluation. 
Once counterfactual consistency holds we can infer that all elements in the counterfactual evidence set equal to those in the factual one, i.e, {$\forall e_j^* \in \textbf{e}^*_\textsc{True}, e_j^* = e_j$, to entail $a^\textbf{R}$ solely. 
In other words, if counterfactual consistency holds, the faithfulness of $a^\textbf{R}$ can be guaranteed with a solid evidence set $\textbf{e}^*$. The counterfactual evaluation score
$S_\text{counter} = 1$ if $a^\textbf{E}_* = a^\textbf{R}$; otherwise $S_\text{counter}$ remains $0$.






So far, the \textit{evaluator} assesses the causal consistency, including non-causal consistency and counterfactual consistency. 
If no causal inconsistency is detected, i.e., $(\forall i^\prime \in \{1, 2, 3\}, g_{i^\prime} = 1)  \wedge (a^\textbf{E}_\text{cf} = a^\textbf{R})$, or $S_\text{non} + S_\text{counter} = 2$ further reasoning may not be required. The \textit{evaluator} agent can confidently accept the original answer, i.e., $a^\textbf{E} = a^\textbf{R}$. 
Otherwise, it proceeds with re-reasoning based on its previous evaluation.
The total evaluation can be written as: 

\begin{small}
\vspace{-4pt}
\begin{equation}
\begin{aligned}
a^\textbf{E} = 
\begin{cases} 
a^\textbf{R} & \text{if } S_\text{non} + S_\text{counter} = 2, \\
\mathrm{Solve}(\textbf{E}, Q, \{g_1, \ldots, g_4\}, \textbf{e}^*) & \text{otherwise}.
\end{cases}
\end{aligned}
\label{eq:9}
\end{equation}
\end{small}
\vspace{-4pt}

Overall, the \textit{evaluator} assesses the causal consistency of $O$, probing for its potential factual accuracy errors and any inference issues and turning out a revised answer $a^\textbf{E}$. 

\subsection{Multi-agent Cooperation for Causal Consistency}
\label{sec:cacocot}

The entire structure of CaCo-CoT is characterized by a recurrent process of reasoning and evaluation for a higher-level faithfulness. 
As discussed in the preceding two sub-sections, the reasoning stage is to elicit \textit{reasoner} agent's reasoning based on human-like causality. The subsequent evaluation stage serves to evaluate top-ranked reasoning chains from \textit{reasoners} based on causal consistency, potentially eliminating factual mistakes and inferential errors.

%
%


In the first round, the algorithm enters the reasoning stage, in which $\{\textbf{R}_1, \ldots, \textbf{R}_{n}\}$ parallelly generate reasoning chain $O_i$ for the question $Q$, where $n$ is the number of \textit{reasoners}. 
The answer implied by $O_i$ is denoted as $a^\textbf{R}_i$. 
If the probability of the top-1 answer in the answer pool, $P(a^\textbf{R}_\text{top1})$, exceeds a threshold $h_0$, such as $50\%$, we assume a consensus among the majority of \textit{reasoner} agents. 
According to whether a consensus between \textit{reasoners} is achieved, the evaluation stage may slightly differ. 

In this stage, if there is a consensus, an \textit{evaluator} agent $\textbf{E}$ examines only a reasoning chain $O_\text{top1}$ randomly sampled from these chains corresponding to the answer $a^\textbf{R}_\text{top1}$. To carry out the counterfactual evaluation, the \textit{evaluator} is provided with an answer $a^* \neq a^\textbf{R}_\text{top1}$ randomly sampled from the answer pool. 
The \textit{evaluator} agent finally returns a modified answer $a^\textbf{E}$. 
If the \textit{evaluator} agrees with $O_\text{top1}$, i.e., $a^\textbf{R}_\text{top1} = a^\textbf{E}$, the algorithm of CaCo-CoT accepts $a^\textbf{E}$ as the final choice. Otherwise, the algorithm proceeds recurrently. 


\begin{algorithm}[h]
\caption{\footnotesize Causal-Consistency Chain-of-Thought (CaCo-CoT)}
\begin{algorithmic}[1]
\begin{scriptsize}

\Procedure{CaCoCoT}{$Q, \mathbb{R}, \mathbb{E}, d, H$}
    \State $h_0, h_1, \hat{d}, k \gets H$ \Comment{Parse hyperparameters}
    
    \State \textcolor{blue}{\textbf{// Reasoning Stage with Multiple Reasoners}}
    \For{$i \in \{1,\ldots,n\}$} \Comment{Parallel reasoning with n reasoners}
        \State $r_1 \gets \mathrm{Expl}(\textbf{R}_i, Q)$ \Comment{Concept explanation, Eq. \ref{eq:2}}
        \State $r_2 \gets \mathrm{Decomp}(\textbf{R}_i, Q, r_1)$ \Comment{Sub-question decomposition, \ref{eq:3}}
        \State $r_3 \gets \emptyset$ 
        \For{$t^\prime \in \{1,\ldots,t\}$} \Comment{Solve each sub-question}
            \State $e_{t^\prime} \gets \mathrm{Solve}(\textbf{R}_i, q_{t^\prime}, (r_1, r_2, \{e_1,\ldots,e_{{t^\prime}-1}\}))$ \Comment{Eq. \ref{eq:4}}
            \State $r_3 \gets r_3 \cup \{e_{t^\prime}\}$
        \EndFor
        \State $a^{\textbf{R}}_i \gets \mathrm{Solve}(\textbf{R}_i, Q, (r_1, r_2, r_3))$ \Comment{Reasoner's answer, Eq. \ref{eq:5}}
        \State $O_i \gets \{r_1, r_2, r_3, a^{\textbf{R}}_i\}$ \Comment{Store complete reasoning chain}
    \EndFor
    
    \State Select the most frequent answer in $\{a^{\textbf{R}}_1,\ldots,a^{\textbf{R}}_n\}$ as $a^{\textbf{R}}_\text{top1}$
    \State $a^{\textbf{R}}_\text{top1}$'s frequency is defined as $P(a^\textbf{R}_\text{top1})$
    
    \State \textcolor{blue}{\textbf{// Evaluation Stage}}
    \If{$P(a^\textbf{R}_\text{top1}) \geq h_0$} \Comment{Strong consensus among reasoners}
        \State $O_\text{top1} \gets \text{RandomSample}(\{O_i \mid a^{\textbf{R}}_i = a^{\textbf{R}}_\text{top1}\})$
        \State $a^* \gets \text{RandomSample}(\{a^{\textbf{R}}_i \mid a^{\textbf{R}}_i \neq a^{\textbf{R}}_\text{top1}\})$

        \State $\textbf{E} \gets \text{RandomSample}(\mathbb{E})$
        \State \textcolor{blue}{\textbf{// Non-causal Evaluation}}
        \For{$i^\prime \in \{1,2,3\}$}
            \State $g_{i^\prime} \gets \mathrm{Eval}(\textbf{E}, Q, O_\text{top1}, (g_1,\ldots,g_{i^\prime-1}))$ \Comment{Eq. \ref{eq:6}}
        \EndFor
        \State $S_\text{non} \gets \frac{1}{3}\sum_{i^\prime=1}^3 g_{i^\prime}$
        
        \State \textcolor{blue}{\textbf{// Counterfactual Evaluation}}
        \State Extract evidence set $\textbf{e}$ from $O_\text{top1}$
        \State $\textbf{e}^*_\text{True} \gets \text{FilterEvidence}(\textbf{e}, Q, \{a^{\textbf{R}}_\text{top1}, a^*\})$ \Comment{Eq. \ref{eq:7}}
        \State $a^{\textbf{E}}_* \gets \mathrm{Solve}(\textbf{E}, Q, \{a^{\textbf{R}}_\text{top1}, a^*\}, \textbf{e}^*_\text{True})$ \Comment{Eq. \ref{eq:8}}
        \State $S_\text{counter} \gets \mathbf{1}[a^{\textbf{E}}_* = a^{\textbf{R}}_\text{top1}]$ \Comment{Indicator: 1 if equal, 0 otherwise}

        \If{$S_\text{non} + S_\text{counter} = 2$}
            \State $a^{\textbf{E}} \gets a^{\textbf{R}}_\text{top1}$ \Comment{Accept original answer, Eq. \ref{eq:9}}
        \Else
            \State $a^{\textbf{E}} \gets \mathrm{Solve}(\textbf{E}, Q, \{g_1,\ldots,g_3\}, \textbf{e}^*_\text{True})$ \Comment{Revise answer}
        \EndIf
        
        \If{$a^{\textbf{E}} = a^{\textbf{R}}_\text{top1}$ \textbf{or} $d \geq \hat{d}$}
            \State \Return $a^{\textbf{E}}$
        \Else
            \State \Return \Call{CaCoCoT}{$Q, \mathbb{R}, \mathbb{E}, d+1, H$}
        \EndIf
        
    \Else \Comment{No strong consensus}
        \State $\mathcal{A}^{\textbf{E}} \gets \emptyset$ \Comment{Evaluator answer pool}
        
        \For{$j \in \{1,\ldots, k\}$} 
            \State Select the $j^\text{th}$ frequent answer $a^{\textbf{R}}_j$
            \State $O_j \gets \text{RandomSample}(\{O_i \mid a^{\textbf{R}}_i = a^{\textbf{R}}_j\})$
            \State $a^* \gets \text{RandomSample}(\{a^{\textbf{R}}_i \mid a^{\textbf{R}}_i \neq a^{\textbf{R}}_j\})$
            \State $\textbf{E}_j$ perform non-causal and counterfactual evaluation and obtain $a^{\textbf{E}}_j$
            \State $\mathcal{A}^{\textbf{E}} \gets \mathcal{A}^{\textbf{E}} \cup \{a^{\textbf{E}}_j\}$
        \EndFor
        
        \State Find most frequent answer in $\mathcal{A}^{\textbf{E}}$ as $a^{\textbf{E}}_\text{top1}$

        \If{$P(a^{\textbf{E}}_\text{top1}) \geq h_1$ \textbf{or} $d \geq \hat{d}$}
            \State \Return $a^{\textbf{E}}_\text{top1}$
        \Else
            \State \Return \Call{CaCoCoT}{$Q, \mathbb{R}, \mathbb{E}, d+1, H$}
        \EndIf
    \EndIf
\EndProcedure

\end{scriptsize}
\end{algorithmic}
\label{alg:1}
\end{algorithm}

Alternatively, if $P(a^\textbf{R}_\text{top1})$ does not surpass the threshold $h_0$, the algorithm selects the top $k$ answers $\{a^\textbf{R}_\text{top1}, \ldots, a^\textbf{R}_{\text{top}k}\}$ from \textit{reasoners}' answer pool, where $k$ is a pre-set integer parameter. 
For each selected answer $a^\textbf{R}_{\text{top}k^\prime} \in \{a^\textbf{R}_{\text{top}1}, \ldots, a^\textbf{R}_{\text{top}k}\}$, we randomly sample $O_{\text{top}k^\prime}$ and $a^*_{\text{top}k^\prime}$ in the same manner as above. 
Next, we initialize $k$ \textit{evaluators} $\{\textbf{E}_1, \ldots, \textbf{E}_k\}$, in which $\textbf{E}_{k^\prime}$ scrutinizes $O_{\text{top}k^\prime}$, considering an extra potential answer $a^*_{\text{top}k^\prime}$, and output $a^\textbf{E}_{k^\prime}$. 
This procedure mirrors the branch described earlier, but it involves a set of top-ranked answers instead of one. 
If these modified answers have an unimodal distribution, i.e., $P(a^\textbf{E}_\text{top1}) > h_1$, the algorithm returns $a^\textbf{E}_\text{top1}$ as the final output. 
However, if no consensus is achieved, the algorithm recursively progresses to the next round of collaborative reasoning until agents reach a consensus or the algorithm hits the wall of the maximum depth $\hat{d}$.

As shown in Alg. \ref{alg:1}, we detail CaCo-CoT using pseudo-code. 
The algorithm takes several key inputs: the question $Q$, a collection of \textit{reasoner} agents $\mathbb{R}$, a collection of \textit{evaluators} agent $\mathbb{E}$, and the recursion depth $d$. The hyper-parameter set $H$ consists of 1) two threshold values $h_0$ and $h_1$; 2) the maximum recursion depth $\hat{d}$ (default: 4); and 3) the number of top-ranked answers $k$ to be evaluated (default: 2).

\section{Experiments}

To evaluate the robustness and effectiveness of our proposed method, a series of experiments are undertaken, comparing it with various existing approaches across a range of benchmarks}. 
In this section, we first compare the performance of CaCo-CoT with state-of-the-art approaches across three text-based knowledge reasoning benchmarks and two multi-modal ones. Then, we move on to the ablation study and case study of CaCo-CoT.

\begin{table*}[h!]
\caption{Comparisons with state-of-the-art approaches on three knowledge-based reasoning datasets based on GPT-3.5-turbo, Claude, and Qwen1.5-32B. The shown metric is accuracy (\%). }
\small
\centering
\resizebox{1.7\columnwidth}{!}{%
\begin{tabular}{c|ccc|ccc|cc}
\toprule[0.8pt]
\multirow{2}{*}{\textbf{Method}} & \multicolumn{3}{c|}{GPT-3.5-turbo} & \multicolumn{3}{c|}{Claude} & \multicolumn{2}{c}{Qwen1.5-32B} \\
      & \textbf{ScienceQA} & \textbf{Com2sense} & \textbf{BoolQ} & \textbf{ScienceQA} & \textbf{Com2sense} & \textbf{BoolQ} & \textbf{ScienceQA} & \textbf{Com2sense} \\ 

\hline
\multicolumn{9}{c}{\textbf{\textit{zero-shot setting}}} \\
\hline                               
Base                                 & 79.3 & 70.1 & 71.7  & 86.8 & 75.3 & 73.4 & 87.4 & 76.0 \\
CoT~\cite{kojima2022large}           & 78.4 & 63.6 & 71.1  & 86.5 & 74.7 & 70.9 & 84.4 & 77.3 \\
SC-CoT~\cite{wang2022self}           & 84.0 & 66.0 & 71.4  & 86.5 & 77.4 & 71.1 & 84.3 & 78.2 \\
C-CoT~\cite{fu2022complexity}        & 82.5 & 68.8 & 70.5  & 86.6 & 76.5 & 70.2 & 84.7 & 78.0 \\
CaCo-CoT (ours)      & \textbf{86.5} & \textbf{73.5} & \textbf{73.5}  & \textbf{89.9}  & \textbf{78.0} & \textbf{76.7} & \textbf{89.8} & \textbf{79.7} \\
\hline
\multicolumn{9}{c}{\textbf{\textit{one-shot setting}}} \\
\hline                                          
CoT~\cite{wei2022chain}            & 82.6 & 70.6 & 70.2    & 89.5  & 78.8 & 72.3 & 89.7 & 78.4 \\
SC-CoT~\cite{wang2022self}         & 86.3 & 72.4 & 70.3    & 90.0  & 80.6 & 70.9 & 90.0 & 78.6 \\
C-CoT~\cite{fu2022complexity}      & 86.9 & 71.9 & 68.9    & 90.1  & 79.7 & 70.7 & 90.1 & 77.7 \\ 
L2M-CoT~\cite{zhou2022least}       & 84.1 & 64.5 & 68.9    & 90.2  & 70.0 & 76.2 & 89.7 & 67.0 \\
BoT~\cite{yang2024buffer}          & 85.3 & 66.2 & 65.7    & 85.8  & 72.5 & 69.1 & 75.5 & 74.6 \\
CaCo-CoT (ours)        & \textbf{88.6} & \textbf{75.0} & \textbf{75.0}  & \textbf{90.8}  & \textbf{83.0} & \textbf{77.9} &\textbf{91.9} & \textbf{80.4} \\
\bottomrule[0.8pt]
\end{tabular}%
}
\vspace{-2pt}
\label{table_1}
\end{table*}

\subsection{Datasets}
\textbf{ScienceQA}~\cite{lu2022learn} is a multiple-choice dataset used to develop a faithful multi-step reasoning model. It consists of approximately 21k questions with rich domain diversity, annotated with image and text context. It is split into train/validation/test sets with a ratio of 60:20:20. In our experiment, we only employ samples from the test set described in the text, with a total number of 2224. We report the accuracy of our method on the set mentioned above.

\textbf{Com2Sense}~\cite{singh2021com2sense} is a commonsense reasoning dataset consisting of 4k sentence pairs annotated with true or false labels. 
It aims to systematically evaluate a model's capability of commonsense knowledge understanding and reasoning. 
%
%
%
In our experiment, the development set is used to benchmark our framework, using accuracy. 

\textbf{BoolQ}~\cite{clark2019boolq} is a reading comprehension dataset labeled with yes/no questions. 
%
%
In detail, it is split into a 3.2k dev set, a 3.2k test set, and a 9.4k train set. 
We use the dev set to evaluate the methods with accuracy, in a closed-book setting. 

\textbf{MME}~\cite{fu2023mme} (Multi-modal Large Language Model Evaluation) benchmark is a comprehensive assessment tool designed to evaluate the capabilities of Multi-modal Large Language Models (MLLMs) across various tasks. MME assesses multi-modal foundation models on a range of abilities, including perception and cognition. Perception tasks involve recognizing specific objects, their existence, count, position, and color, while cognition tasks include commonsense reasoning, numerical calculation, text translation, and code reasoning. To match the goal of our approach, we evaluate the multi-modal reasoning capability of our approach on the cognition split only.

\textbf{MMMU}~\cite{yue2023mmmu} (Massive Multi-discipline Multi-modal Understanding) stands out as a rigorous benchmark that specifically challenges AI models' advanced cognitive capabilities and expert-level reasoning. MMMU features 11,500 college-level questions across six academic disciplines, 30 subjects, and 183 subfields, incorporating 32 different types of visual elements from charts to musical notation. MMMU successfully identifies areas where current AI technology needs improvement, particularly in handling complex, discipline-specific problems. While the most advanced GPT-4V~\cite{openai2023gpt4} achieves only 56\% accuracy, MMMU successfully identifies areas where current AI technology needs improvement, particularly in handling complex, discipline-specific problems.

\subsection{Experiment Details}

\label{ap:exp_details}
In the following experiments, we evaluate CaCo-CoT based on a spectrum of foundation models, including large language models (LLMs) and MLLMs across different parameter sizes and architectures. LLMs involve GPT-3.5-turbo~\cite{ouyang2022training}, Claude-2~\cite{anthropic2023modelcardevaluation}, Qwen1.5-32B~\cite{bai2023qwen}, while MLLMs are Llama3-LLaVA-Next-8B~\cite{liu2024llavanext}, InternVL2-8B~\cite{chen2023internvl}, Qwen-VL-Max~\cite{Qwen2VL}, GPT-4o-mini, and GPT-4o~\cite{openai2023gpt4}.

We invoke official APIs provided by OpenAI, Anthropic, and Alibaba Cloud to obtain the corresponding models' results, except for InternVL2-8B and Llama3-LLaVA-Next-8B, which are deployed offline. The parameter $top\_p$ is typically set to 0.4, and $temperature$ remains 0.5 for single sampling approaches. 
For self-consistency CoT~\cite{wang2022self} (or SC-CoT) that samples 10 responses for each query, higher $top\_p$ and $temperature$ of 0.5 and 0.8 are applied for the diversity of reasoning chains. 
We notice that contemporary foundation models have ingested a vast amount of knowledge, and their reasoning capability is not constrained by a lack of knowledge and their inability to mimic human reasoning structures. 
%
%
For the one-shot setting in Table ~\ref{table_1}, we share a demonstration question from the training set for all approaches. 
The demonstration reasoning chains for different approaches are generated by GPT-4~\cite{openai2023gpt4}. 

\subsection{Comparisons with State-of-the-art Methods}

In this section, we make comparisons between CaCo-CoT and its counterparts on two tasks, science question answering and commonsense reasoning purely based on text, as well as multi-modal knowledge reasoning. To more concisely refer to methods, we shorten Chain-of-thought~\cite{kojima2022large} to CoT, Self-consistency CoT~\cite{wang2022self} to SC-CoT, Complexity-based CoT~\cite{fu2022complexity} to C-CoT, Least-to-Most~\cite{zhou2022least} to L2M-CoT, Buffer-of-thoughts to BoT~\cite{yang2024buffer} and Compositional CoT~\cite{MitraCCoT} to CompCoT.

\subsubsection{Science Question Answering} Science question answering embodies the procedure of memorizing and employing scientific knowledge to comprehend, explain, and predict phenomena in the natural world. 
Table \ref{table_1} compares our method with several methods to elicit reasoning in foundation models, including CoT, SC-CoT, C-CoT, and L2M-CoT in both zero-shot and one-shot settings. 
Here, L2M-CoT can only be implemented in a one-shot manner as it has to demonstrate how to decompose problems and address sub-questions one by one. Similarly, BoT retrieves relevant solutions to the input question from an online maintained buffer, which is deemed a one-shot setting in this paper. 
In Table \ref{table_1}, CaCo-CoT's implementation comprises only two \textit{reasoners}, and the maximum number of \textit{evaluators} $k$ is set to $2$.

As shown in Table \ref{table_1}, CaCo-CoT outperforms the existing methods using three different foundation models, GPT-3.5-turbo~\cite{ouyang2022training}, Claude~\cite{bai2022constitutional}, and an open-source LLM Qwen1.5-32B~\cite{bai2023qwen} in both zero-shot and one-shot settings. 
We also find that CaCo-CoT's zero-shot performance slightly outperforms the one-shot performance of SC-CoT and C-CoT on BoolQ. 
This demonstrates the superiority of our proposed CaCo-CoT, in terms of recalling factual knowledge and reducing inferential fallacies. 



\subsubsection{Commonsense Reasoning} 
Commonsense reasoning aims to answer questions that require reasoning over general background knowledge. Commonsense reasoning capability is essential for artificial intelligence deployed in the real world. Strong commonsense reasoning capabilities confine it to functioning within reasonable and safe boundaries.


\begin{table*}[t]
\small
\centering
\caption{Comparison of different cooperation modes on the textual split of ScienceQA. NAT, SOC, and LAN refer to natural science, language science, and social science respectively. The level of grade represents the difficulty of the question. \,\dag\, denotes one-shot settings, while its absence indicates zero-shot settings. $k$ is the maximum number of \textit{evaluators} to be invoked in each round. } 
\resizebox{1.75\columnwidth}{!}{%
\addtolength{\tabcolsep}{4pt}
\begin{tabular}{c|c|ccc|cc|c}
\toprule[1pt]
\multirow{2}{*}{Method} & \multirow{2}{*}{\parbox{2cm}{\centering Category}} & \multicolumn{3}{c|}{Subject} & \multicolumn{2}{c|}{Grade} & \multirow{2}{*}{Average} \\ 
     &  & NAT & SOC & LAN & G1-6 & G7-12 &  \\ \hline 

CoT${}$~\cite{kojima2022large}       & \multirow{5}{*}{Single-chain}       & 85.71 & 86.40 & 77.37  & 82.70 & 80.52 & 81.79 \\
CoT${\dag}$~\cite{wei2022chain}    &      & 85.91 & 92.00 & 82.95 & 86.41 & 82.67 & 84.85 \\ 
L2M-CoT~\cite{zhou2022least} & & 83.41 & 91.20 & 83.89  & 86.17 & 81.16 & 84.08 \\
CaCo-CoT${}_{\mathrm{R} 1 } \; \textbf{w/o} \; \textit{evaluator}$ &  & 85.52 & 95.20 & 82.01 & 86.02 & 82.13 & 84.40  \\
CaCo-CoT${\dag}_{\mathrm{R} 1}  \; \textbf{w/o} \; \textit{evaluator}$  & & 89.26 & 96.00 & 81.44 & 88.34 & 82.56 & 85.93 \\ \hline 

SoT~\cite{ning2024skeletonofthought}  &\multirow{1}{*}{\parbox{2cm}{\centering Ensembling}}   & 64.72 & 73.60 & 69.41 & 70.73 & 62.86 & 67.45 \\ 
\hline

SC-CoT${\dag}$  ~\cite{wang2022self}  &\multirow{4}{*}{\parbox{2cm}{\centering Majority voting}}   & 88.78 & 91.20 & 83.33 & 88.57 & 83.21 & 86.33 \\
C-CoT${\dag}$ ~\cite{fu2022complexity} &  & 88.78 & 91.20 & 84.47 & 89.03 & 83.85 & 86.87 \\ 

CaCo-CoT${}_{\mathrm{R} 3} \; \textbf{w/o} \; \textit{evaluator} $ & & 87.44 & 92.00 & 83.43 & 87.49 & 83.42 & 85.79 \\
CaCo-CoT${\dag}_{\mathrm{R} 3} \; \textbf{w/o} \; \textit{evaluator}$ &  & 90.79 & \textbf{99.20} & 83.32 & 89.79 & 84.82 & 87.71 \\[1pt] \hline

CaCo-CoT${}_{\mathrm{R} 2} \;(p=2)$ &\multirow{4}{*}{\parbox{2cm}{\centering Multi-round cooperation}}  & 88.78 & 93.60 & 83.43 & 88.57 & 83.64 & 86.51 \\
CaCo-CoT${\dag}_{\mathrm{R} 2} \;(p=2)$ &  & 91.37 & 97.60 & \textbf{84.19} & 90.12 & 85.79 & 88.31 \\
CaCo-CoT${}_{\mathrm{R} 3} \;(p=2)$ &  & 89.45 & 96.00 & 83.52 & 89.03 & 84.18 & 87.01 \\

CaCo-CoT${\dag}_{\mathrm{R} 3} \;(p=2)$ &
& \textbf{93.38} & 96.80 & 83.62 & \textbf{90.97} & \textbf{86.11} & \textbf{88.94}  \\

\bottomrule[1pt]
\end{tabular}%
}

\label{table2}
\end{table*}

Our method outperforms the existing methods in both Com2Sense and BoolQ, as displayed in Table \ref{table_1}. 
As we observed, CoT-based approaches~\cite{wei2022chain, wang2022self} usually fall short of employing relevant background knowledge implied by the question, because they often embark on knowledge inference before having a clear definition of the knowledge implied by the question. 
Notably, they usually proceed in a "plan-as-you-go" manner (without a clear plan to follow), which may be misleading for multi-hop reasoning problems. 
While BoT~\cite{yang2024buffer} demonstrates exceptional performance in puzzle games, such as Game24 and Tower of Hanoi, through the use of a code interpreter, it does not exhibit comparable superiority in knowledge-based reasoning tasks. This limitation arises because BoT struggles to sustain the high quality of solutions stored in its buffer, potentially leading to a flawed or misleading reasoning chain when addressing input questions.

%
In contrast, CaCo-CoT employs relatively a more faithful and effective way to perform knowledge inference, that is to explicitly extract the knowledge and decompose the question into a group of manageable sub-questions before reasoning. 
Moreover, \textit{evaluators} inspects the reasoning chain for causal consistency, searching for solid evidence for its potential answer. 
As shown, there is more than 3\% accuracy improvement in the zero-shot setting and 2.6\% in the one-shot setting on Com2sense in GPT-3.5-turbo. 
Meanwhile, there is a noticeable accuracy improvement of approximately 3.3\% on BoolQ and 2.4\% on Com2sense when using Claude. Additionally, CaCo-CoT based on the open-source Qwen1.5-32B outperforms other approaches by 2.4\% on ScienceQA.
Overall, CaCo-CoT showcases its powerful knowledge reasoning capability by outperforming state-of-the-art counterparts on three benchmarks based on open-source or closed-source foundation models.

\subsubsection{Multi-modal knowledge reasoning}



We extend CaCo-CoT into the realm of multi-modal knowledge reasoning, to further ascertain the limits of its capability. 

\begin{figure}[t!]
\centering
\includegraphics[width=0.49\textwidth]{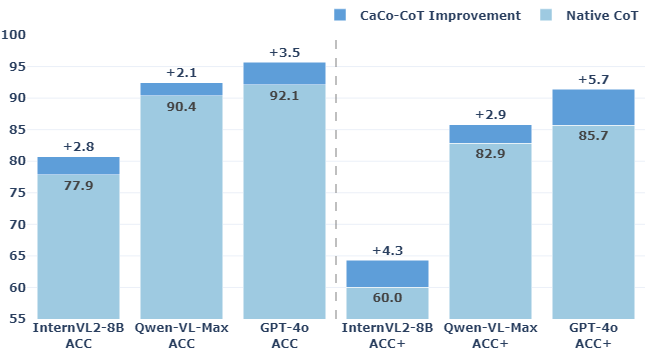}
\caption{Performance Improvement of CaCo-CoT over Native CoT on the MME commonsense reasoning split. Following the standard setting~\cite{fu2023mme}, ACC (\%) calculates the accuracy of each image-question pair. ACC+ measures if two questions associated with an image are both answered correctly.  }
\label{fig:mme}
\end{figure}

As shown in Fig. \ref{fig:mme}, we first evaluate the performance of CaCo-CoT on the MME commonsense reasoning split against the baseline, native CoT that comes with foundation models themselves. The experimental results demonstrate that our proposed CaCo-CoT consistently outperforms the native CoT approach across all three MLLMs tested, with improvements ranging from 1.39\% to 3.77\%. Specifically, using CaCo-CoT with GPT-4o achieved the best performance, reaching 95.68\% ACC and 91.42\% ACC+, representing improvements of 3.54 and 5.71\% respectively over the native CoT baseline. Similar performance gains are observed with InternVL2-8B, where CaCo-CoT improved ACC by 2.85\% (from 77.86\% to 80.71\%) and ACC+ by 4.29\% (from 60.00\% to 64.29\%). These results suggest that CaCo-CoT provides robust enhancement across different model architectures and scales, with the more substantial improvements in ACC+ indicating stronger reasoning consistency when answering multiple questions about the same image. 


On the more challenging MMMU benchmark, experimental results demonstrate the superior performance of CaCo-CoT across different model scales. Using Llama3-LLaVA-Next-8B~\cite{liu2024llavanext} as the base model, CaCo-CoT achieves 48.21\% accuracy on the challenging MMMU benchmark, notably outperforming other methods including Native CoT (37.53\%), CompCOT~\cite{MitraCCoT} (40.52\%), and ReAct~\cite{yao2023react} (46.70\%). This improvement highlights CaCo-CoT's effectiveness in enhancing multi-modal reasoning capabilities. The performance advantage persists with powerful models, where CaCo-CoT reaches 60.09\% accuracy with GPT-4o-mini~\cite{openai2023gpt4}, maintaining its lead over competing methods.

\begin{figure}[t!]
\centering
\includegraphics[width=0.49\textwidth]{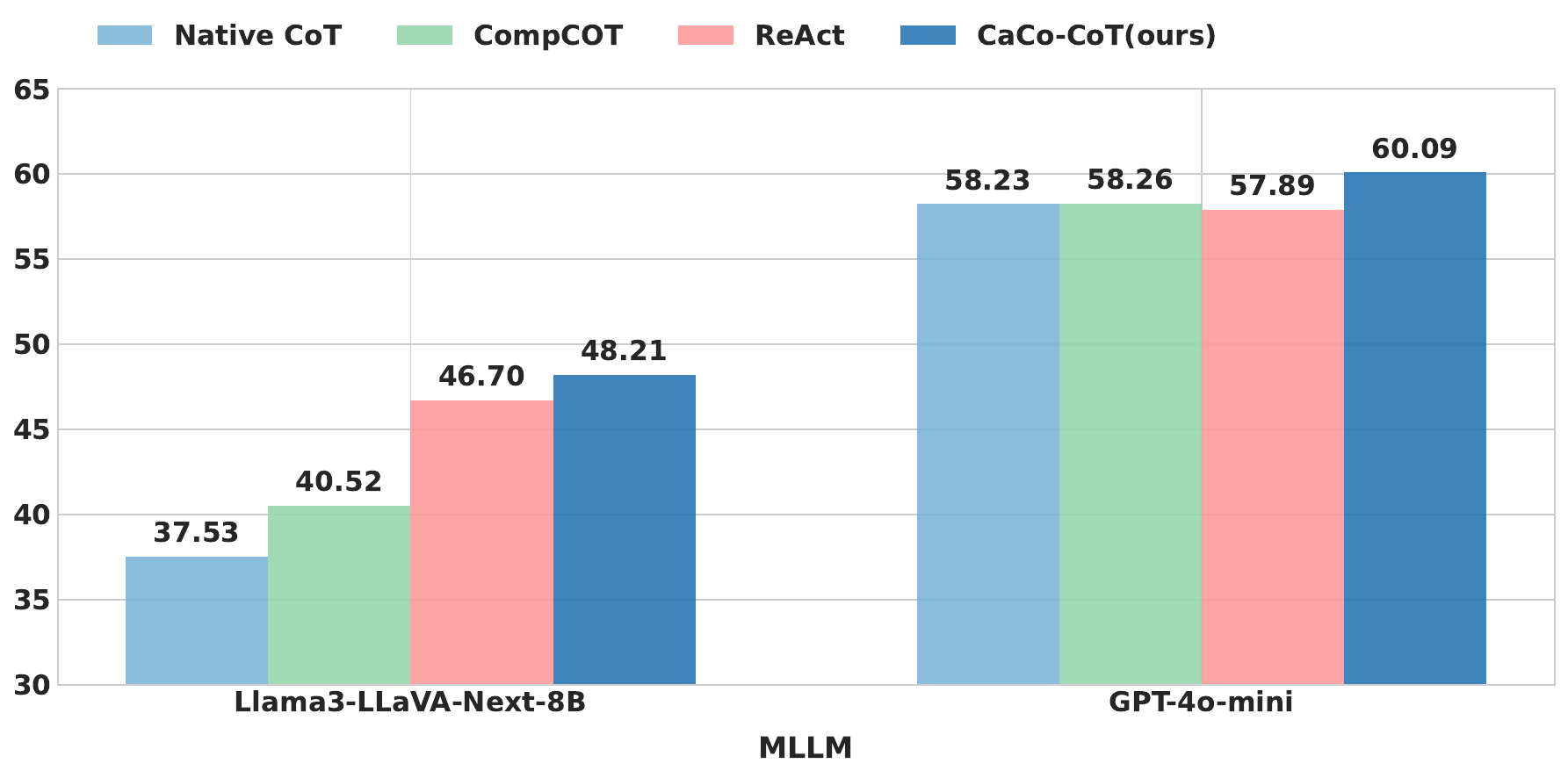}
\vspace{-20pt}
\caption{Performance comparison on the MMMU dataset with Llama3-LLaVA-Next-8B and GPT-4o-mini. }
\label{fig:mmmu}
\end{figure}





\subsection{Ablation Study}
%

\begin{figure*}[t]
\vspace{-5pt}
\centering
\includegraphics[width=0.98\textwidth]{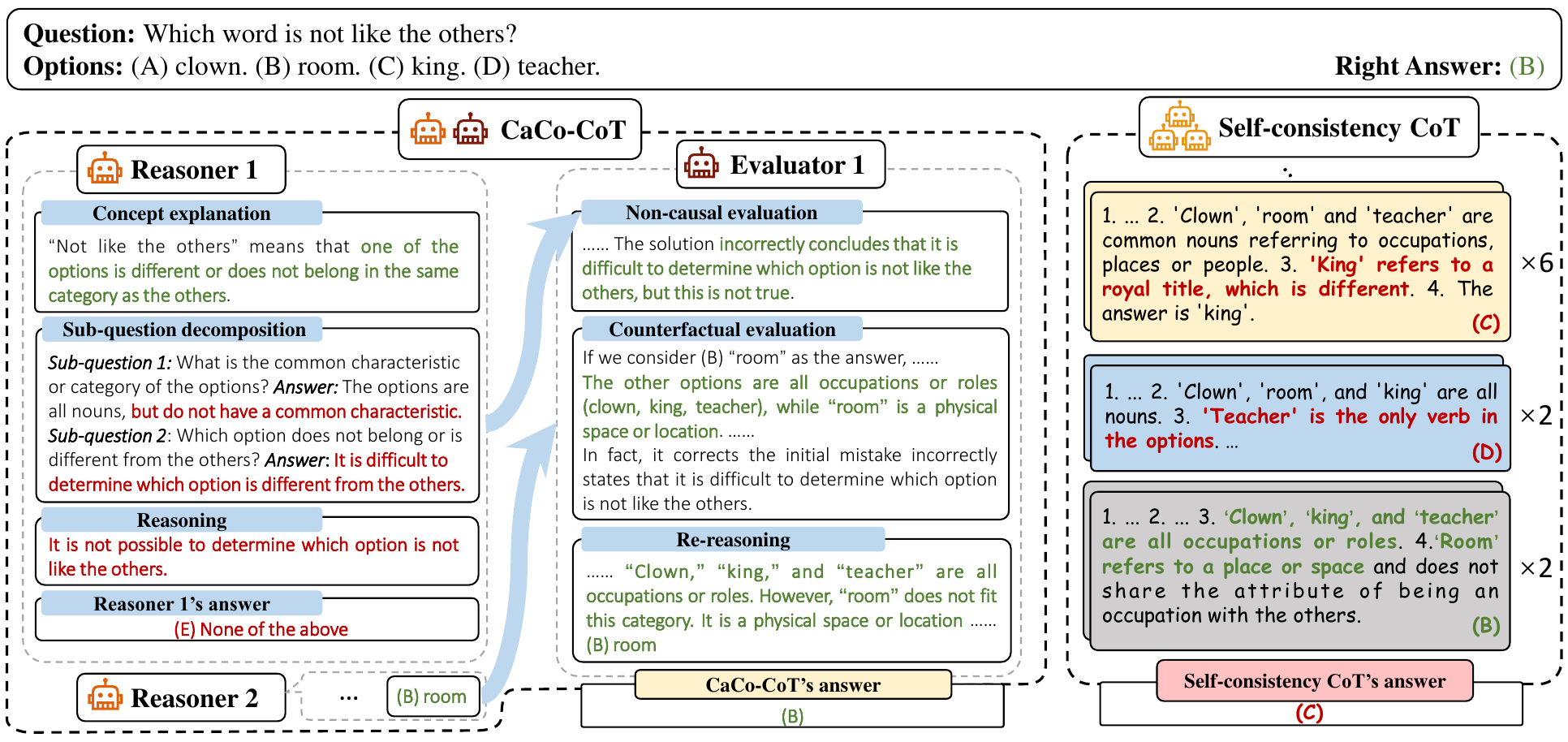}
\vspace{-10pt}
\caption{Zero-shot CaCo-CoT versus zero-shot SC-CoT in resolving a question on ScienceQA. Text highlighted in green indicates positive contents that contribute to the correct answer, while the red portions represent unfaithful or misleading content. } 
\label{fig:case}
\vspace{-10pt}
\end{figure*}

\subsubsection{Agent component}

To study the individual components proposed in this paper, we conduct a range of ablative experiments on the Com2sense dataset. 
As shown in Table~\ref{table3}, we first ablate the \textit{concept explanation} component of the \textit{reasoner}. 
After that, the \textit{non-causal evaluation} component and \textit{counterfactual evaluation} component of the \textit{evaluator} are respectively ablated.
The resulting accuracy reveals that the causal-consistency-seeking process is crucial for the commonsense reasoning task. When the \textit{concept explanation} component is employed, performance balloons, with gains of up to 3.9\% on GPT-3.5-turbo and 3.8\% on Claude. 
The \textit{non-causal evaluation} component that examines the reasoning chain after a full scan of it increases the final accuracy by 3.3\% on GPT-3.5-turbo and 1.7\% on Claude. 
Moreover, without inspecting reasoning chains through a counterfactual reasoning process, termed \textit{counterfactual evaluation}, both models exhibit a dramatic drop in accuracy by 1.5\% and 1.0\% respectively.

\begin{table}[t!]
\footnotesize
\centering
\caption{Ablation Study of CaCo-CoT's components on Com2sense~\cite{singh2021com2sense}.} 
\begin{tabular}{c|c|c} 
\toprule[1pt]

Model      &  Method      & Accuracy  \\ \hline

\multirow{5}{*}{GPT-3.5-turbo}    & CaCo-CoT \textbf{w/o} \textit{concept explanation}                     & 69.6  \\ 
                                  &  CaCo-CoT \textbf{w/o} \textit{non-causal evaluation}      & 70.2  \\
                                  & CaCo-CoT \textbf{w/o} \textit{counterfactual evaluation}  & 72.0  \\
                                 & CaCo-CoT  & \textbf{73.5}      \\  \hline
\multirow{4}{*}{Claude}           & CaCo-CoT \textbf{w/o}  \textit{concept explanation}                     & 74.1  \\ 

                                  & CaCo-CoT \textbf{w/o} \textit{non-causal evaluation}      & 76.2  \\
                                  &  CaCo-CoT \textbf{w/o}  \textit{counterfactual evaluation}  & 76.9  \\
                                  &  CaCo-CoT          & \textbf{78.0}  \\

\bottomrule[1pt]
\end{tabular}%
\vspace{-7pt}
\label{table3}
\end{table}

\subsubsection{Multi-agent cooperation}

\begin{figure*}[h!]
\centering
\vspace{-10pt}
\includegraphics[width=0.98\textwidth]{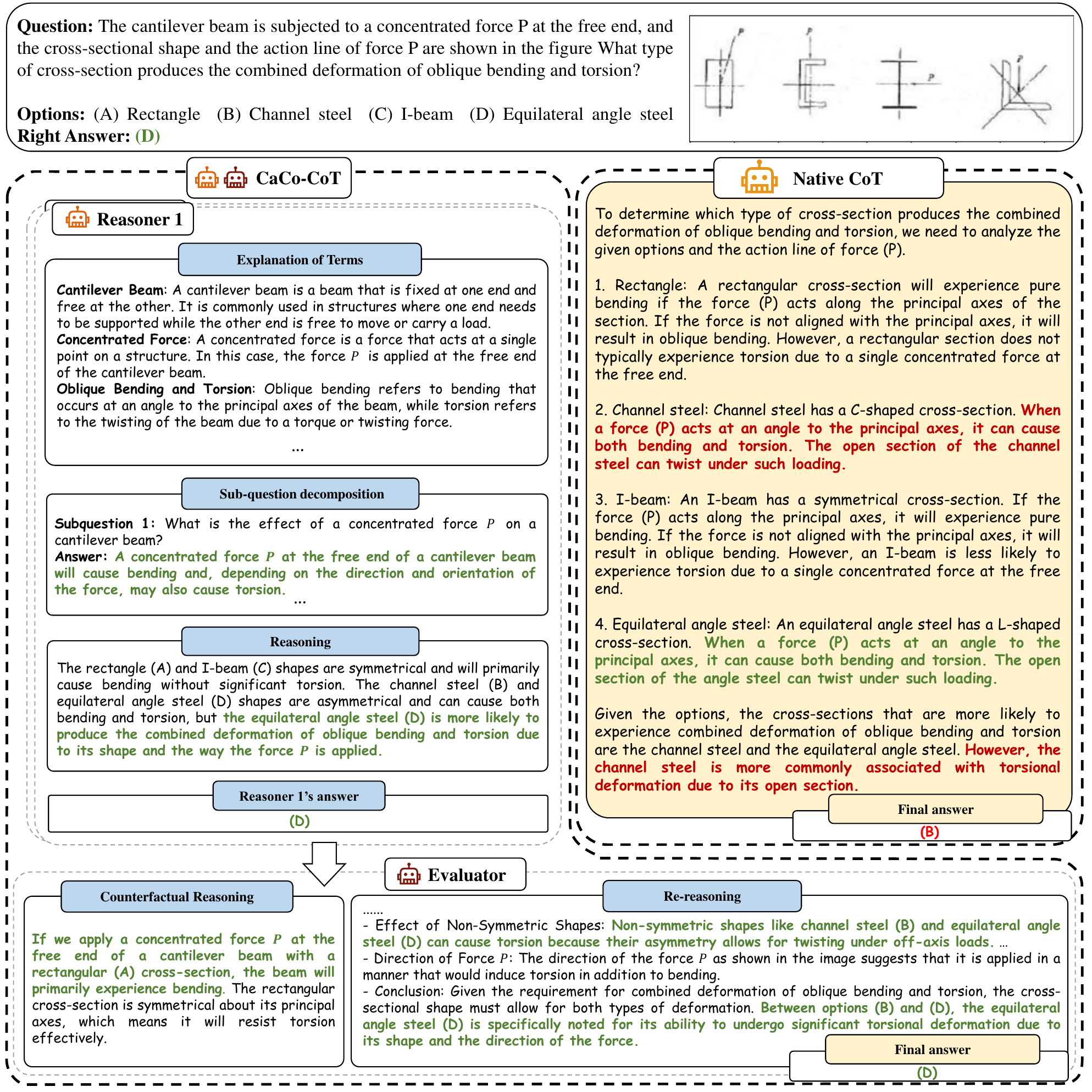}
\caption{A case study demonstrating multi-modal reasoning in a structural engineering analysis problem on MMMU. The example shows how CaCo-CoT and Qwen-VL-Max's native CoT leverage multi-agent cooperation to analyze beam deformation, with green text highlighting correct analytical steps and red text indicating potentially misleading reasoning or conflicted content. The case illustrates how CaCo-CoT handles complex interplay between visual analysis and textual problem-solving in determining combined oblique bending and torsional deformation.}
\label{fig:mmcase}
\vspace{-5pt}
\end{figure*}

Table \ref{table2} presents a comparison of the results obtained from various implementations of the proposed method, CaCo-CoT, alongside some compared methods. These implementations are evaluated across diverse disciplines and levels on the ScienceQA dataset.
For SC-CoT and C-CoT, we sample 10 reasoning chains for majority voting, following the original papers. SoT is short for Skeleton-of-Thought~\cite{ning2024skeletonofthought}  that creates a structured outline (skeleton) of the answer and then fills in the details through parallel processing. 
For our variance of voting among three \textit{reasoners}, if two or more than two \textit{reasoners} arrive at a single answer, we accept the answer as the final output.

\begin{figure*}[t!]
\centering
\vspace{-20pt}
\includegraphics[width=0.98\textwidth]{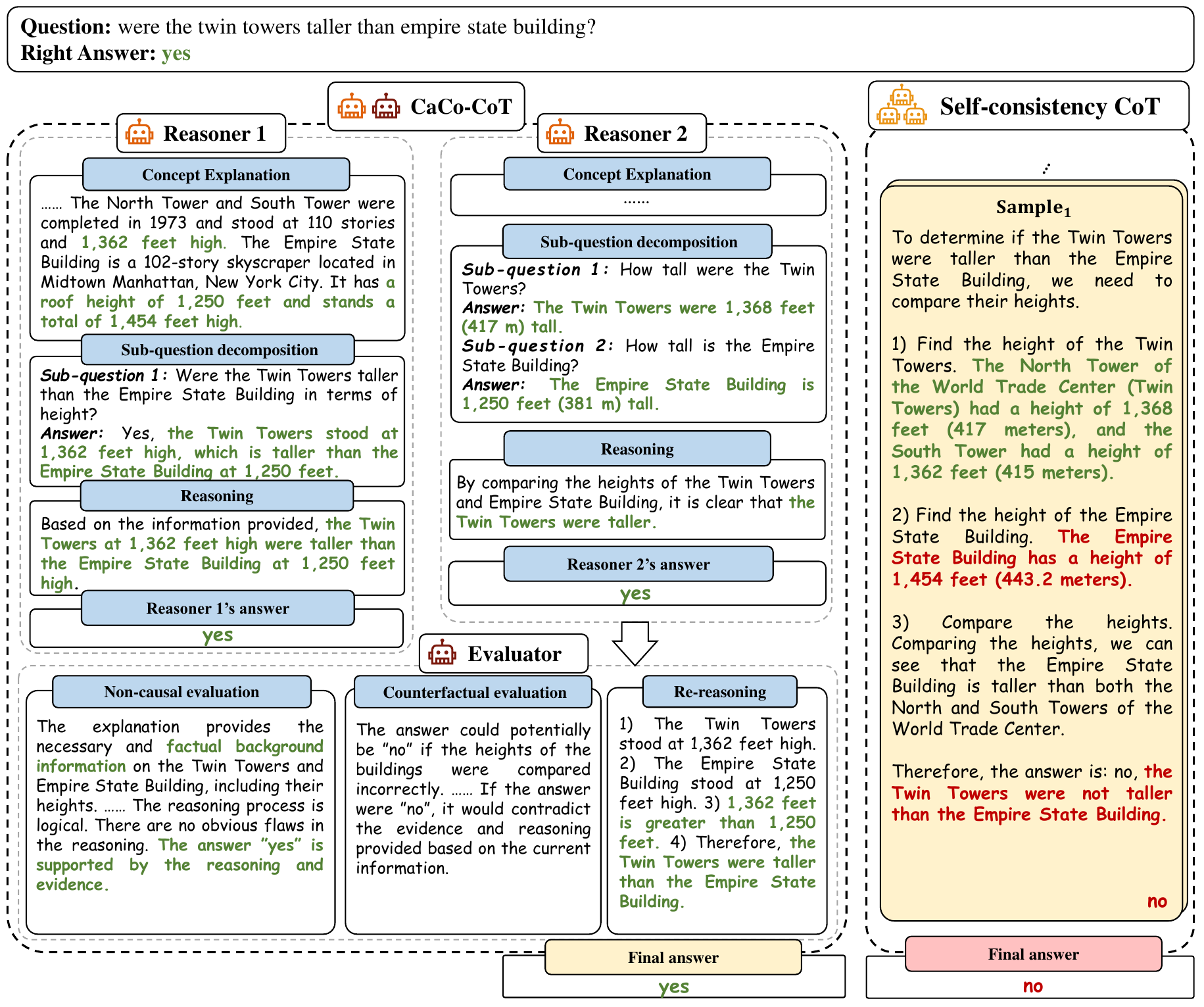}
\caption{A single-round example of CaCo-CoT$\dag$ versus SC-CoT$\dag$ on a BoolQ dataset question about building heights comparison, where $\dag$ denotes the one-shot settings. The green text indicates the positive content that contributes to the correct answer, while the red portions represent unfaithful or misleading content.}
\vspace{-10pt}
\label{fig:7}
\end{figure*}

Initially, we focus on the effectiveness of the \textit{reasoner}. Compared to single-sampling counterparts on both settings, our single \textit{reasoner}, i.e., CaCo-CoT$_{\mathrm{R} 1}$, showcases its superior knowledge reasoning power due to the human-like causal instruction. 
With components of \textit{concept explanation} and \textit{sub-question decomposition}, foundation models are prompted to self-retrieve their internal knowledge that may shed light on how they perform knowledge inference. 
We empirically find that, even though in our one-shot CoT reproduction, the one-shot demonstration explains the concept in the first sentence, foundation models may fail to generalize to questions of different formulations or different disciplines. 
In comparison with other majority voting-based approaches~\cite{wang2022self, fu2022complexity} in the one-shot setting, the majority voting among only three \textit{reasoners} surpasses by a 0.9\% margin in average accuracy, while achieving 99.2\% accuracy in social science problems.

Aimed to correct factual and inferential mistakes through causal consistency evaluation, we introduce the \textit{evaluator}. 
Comparing the results of "CaCo-CoT$_{\mathrm{R} 2}$" with that of "CaCo-CoT$_{\mathrm{R} 3} \; \textbf{w/o} \; \textit{evaluator}$" reveals that the performance gain does not come from an increase in the number of agents but the improvement of causal consistency brought by the \textit{evaluator}. 
When the number of \textit{reasoners} grows, the performance in both zero-shot and one-shot settings balloons without reaching a ceiling. 
Furthermore, the experiments of CaCo-CoT$_{\mathrm{R} 3}$ and CaCo-CoT$_{\mathrm{R} 3} \; \textbf{w/o} \; \textit{evaluator}$ in both settings discover that even if three independent \textit{reasoners} produce erroneous inference and incorrect answers, the \textit{evaluator} is can correct them, with average accuracy lifted by over 1.2\%. 
These experiments demonstrate the effectiveness of our proposed causal evaluator agent. 
Notably, our implementation of CaCo-CoT${\dag}_{\mathrm{R} 3}$ dramatically boosts the accuracy in harder questions of grade 7-12 to 86.11\%, only taking a grade-2 demonstration, while previous approaches reach a limit of 83.85\%.



\subsection{Case Study}
\label{sec:ap_case}

In this section, we demonstrate our approach's pros and cons through several case studies.


%
As illustrated in Fig. \ref{fig:case}, to mitigate the uncertainty of SC-CoT, we sample ten responses from GPT-3.5-turbo to the same query and display the most representative responses from each category. 
We observe that it occasionally provides the correct answer (the light green box Fig. \ref{fig:case}), albeit with a low probability. 
But more frequently, it makes typical mistakes such as lacking timeliness in understanding the question (the second light yellow box in Fig. \ref{fig:case}); inference errors (the top light yellow box); and false knowledge (the light blue box). These phenomena indicate that if users aim to obtain the correct answer, they should repeatedly inquire with foundation models, which is highly inefficient. 
In our method, even though the sub-step of \textit{concept explanation} clarifies the intent of the question, the \textit{reasoner} is confused during the problem decomposition process and refuses to provide a discriminative answer. 
However, the \textit{evaluator} 1) identifies mistakes made by the \textit{reasoner}, e.g., confusion in the sub-question decomposition step; 2) incorporates factual evidence to support the right answer, i.e., ``\textit{the other options are all occupations or roles (clown, king, teacher), while `room' is a physical space or location}''. 3) Finally, it corrects the final choice to the golden one, (B) room.

\subsubsection{Multi-modal Reasoning}
Fig. \ref{fig:mmcase} illustrates how CaCo-CoT attempts to answer a structural engineering analysis problem on MMMU. CaCo-CoT demonstrates superior problem-solving capabilities compared to native CoT in analyzing the cantilever beam question through its comprehensive, multi-layered approach. While native CoT simply lists properties of each cross-section shape and arrives at an incorrect conclusion (B) without systematic verification, CaCo-CoT breaks down the problem into fundamental sub-questions about force effects and cross-sectional properties. It employs multiple \textit{reasoners} providing different perspectives, explicitly considers a counterfactual scenario by examining why certain shapes would not work (particularly the rectangular shape), and includes clear explanations of key points to figure out the question. This structured approach enables CaCo-CoT to achieve the correct answer (D) by evaluating the specific conditions needed for combined deformation. Furthermore, CaCo-CoT's reasoning helps identify and correct potential misconceptions that might arise from native CoT, particularly in understanding how the direction of force $P$ relates to each shape's behavior. Its verification process effectively prevents the kind of errors seen in the native CoT approach, which failed to catch its misconception about channel steel's torsional deformation properties.

\subsubsection{Faithful Reasoning through Improving Factual Accuracy}

In Fig. \ref{fig:7}, we showcase how our approach significantly elicits more faithful knowledge reasoning in foundation models by lowering factual errors. In this case, CaCo-CoT's \textit{reasoners} provide accurate and factual background knowledge about the heights of the Twin Towers and the Empire State Building. By simply comparing them, our \textit{reasoners} derive the correct answer \textit{yes}. By comparison, a one-shot sample (\textbf{Sample}$_1$ in SC-CoT offers a hallucination, i.e., the heights of two buildings are both in factually inaccurate numbers, which directly leads to a mistake in the final answer. Note that the base model behind these two approaches is identical. We only prompt the \textit{reasoners} to recall the background knowledge in the first place, which is straightforward and efficient.

The limitations of contemporary foundation models in handling dictionary guide word questions are related to the underlying word embedding technique used in these models. It focuses more on capturing semantic relationships among words based on their contextual usage patterns in the training data instead of word structure information found in dictionaries. Eventually, words with similar meanings or usage contexts are represented by high-dimensional vectors that are closer together in this embedding space. As a result, foundation models may struggle to provide concise answers for this type of query. This may be one of the intrinsic biases within foundation models which we cannot fully address in this paper.

\begin{figure}[h!]
\centering
\includegraphics[width=0.9\columnwidth]{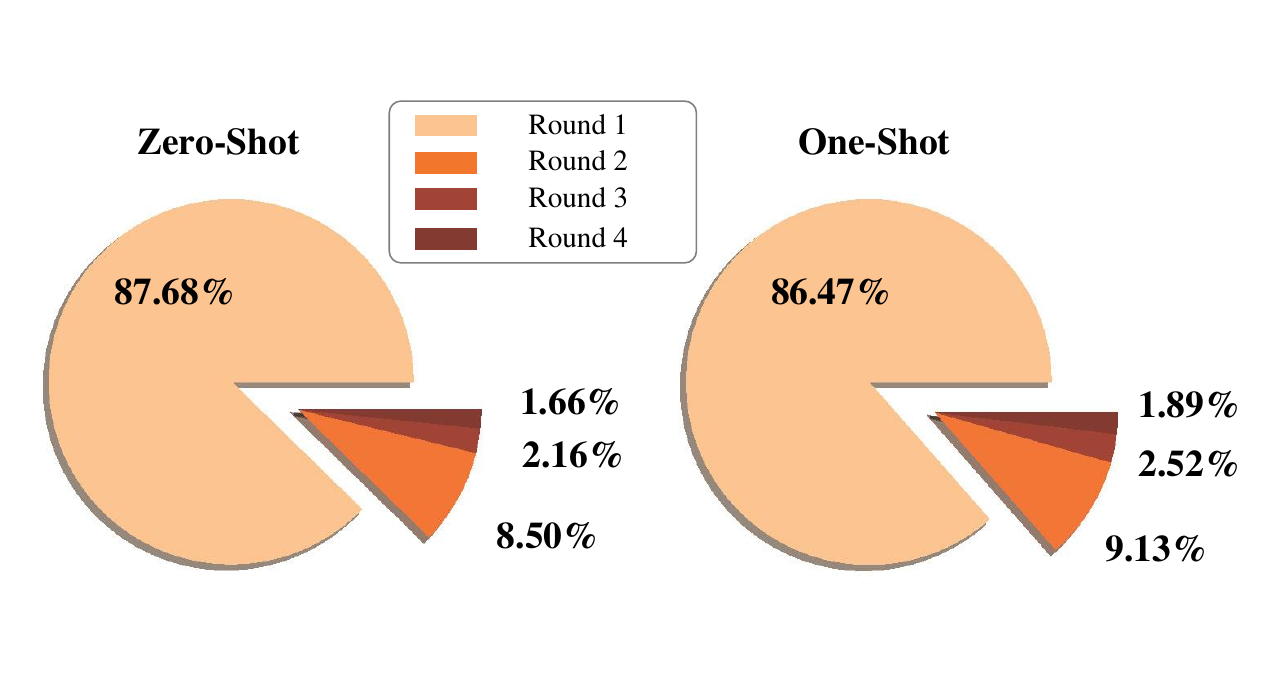}
\vspace{-20pt}
\caption{The distribution of cooperation rounds to reach a consensus. }
\vspace{-10pt}
\label{fig:depth}
\end{figure}

\subsection{Efficiency Analysis}
To evaluate the effectiveness of CaCo-CoT, we statistically analyze the number of cooperation rounds in both zero-shot and one-shot settings using the ScienceQA dataset. We set the maximum depth $\hat{d}$ to 4. The findings are depicted in Fig. \ref{fig:depth}. 
%
Remarkably, cooperation beyond one round accounts for less than 14\% of cases, with over 86\% completing in a single round. Regarding time complexity, while agents can be invoked concurrently in each phase, the time complexity for either reasoning or evaluation phases is $O(1)$. Consequently, the expected time complexity of CaCo-CoT is effectively linear, with a coefficient that reflects the average number of rounds needed multiplied by 2 (reasoning and evaluation phases in each round). Here, the expected time consumption is merely 2.356 times that of each agent invocation.

\section{Conclusion}


To tackle the significant challenge of factual errors and inferential fallacies inherent in contemporary foundation models, we present a novel framework called CaCo-CoT, inspired by the principles of multi-agent cooperation. CaCo-CoT incorporates two distinct types of agents, namely \textit{reasoner} and \textit{evaluator}, collaborating within a reasoning-and-consensus paradigm to achieve reasoning chains that adhere to causal consistency. Through extensive experiments spanning various knowledge reasoning tasks, CaCo-CoT demonstrates faithful results and state-of-the-art performance, highlighting the effectiveness of improving faithfulness and causality.

\section{Limitations}
\label{ap:limitations}
Concerning the limitations of our approach, when the collection of agents share a foundation model, they harbor shared biases inevitably. This is embodied when they confidently reason out an erroneous outcome. Also, individual \textit{reasoners} occasionally exhibit error accumulation, wherein errors in the prerequisite steps cause more errors in subsequent steps. Especially, \textit{reasoners} can adapt to the golden answer when the background knowledge (golden evidence) is given. In addition, our method is unable to effectively screen out the most faithful solution under extreme circumstances (less than 1\%), such as continuous inconsistencies.

\clearpage
\bibliography{reference}
\bibliographystyle{IEEEtran}












\vspace{-16.5pt}
\begin{IEEEbiography}[{\includegraphics[width=1in,height=1.25in,clip,keepaspectratio]{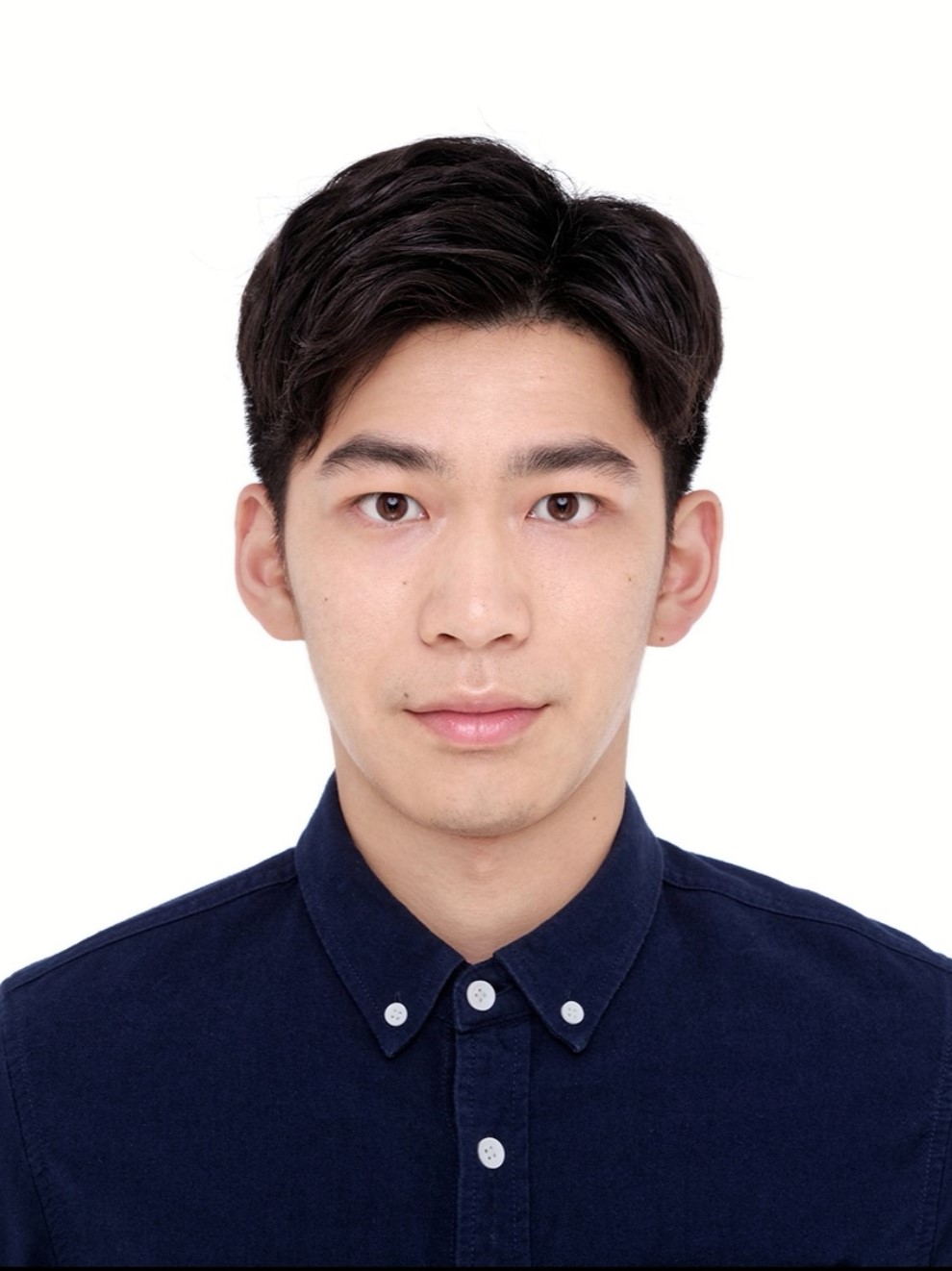}}]{Ziyi Tang} is now pursuing his Ph.D. degree at Sun Yat-Sen University. Prior to this, he was a research assistant at The Chinese University of Hong Kong, Shenzhen, China. He received his B.E. degree from South China Agriculture University, Guangzhou, China in 2019 and M.S. degree from The University of Southampton, Southampton, U.K. in 2020. He has achieved top places in data science competitions hosted by Kaggle and Huawei respectively. He has authored and co-authored 3 papers in top-tier academic journals and conferences. He has also served as a reviewer for ICLR and TOMM. His research interests include multi-modal reasoning, large language models, and causal reasoning. 
\end{IEEEbiography}

\vspace{-16.5pt}
\begin{IEEEbiography}[{\includegraphics[width=1in,height=1.25in,clip,keepaspectratio]{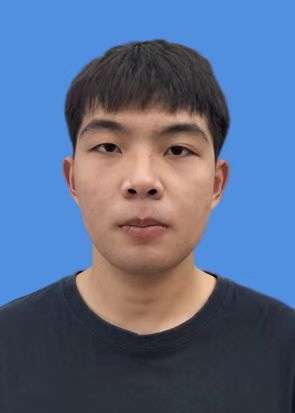}}]{Ruilin Wang} has received the B.S. degree from the School of Computer Science and Engineering, Northeastern University in 2023. He is now pursuing his M.S. degree at the School of Computer Science and Engineering, Sun Yat-sen University. His main interests include motion generation, image editing, and language model reasoning. He has been serving as a TVCJ reviewer.
\end{IEEEbiography}

\vspace{-16.5pt}
\begin{IEEEbiography}[{\includegraphics[width=1in,height=1.25in,clip,keepaspectratio]{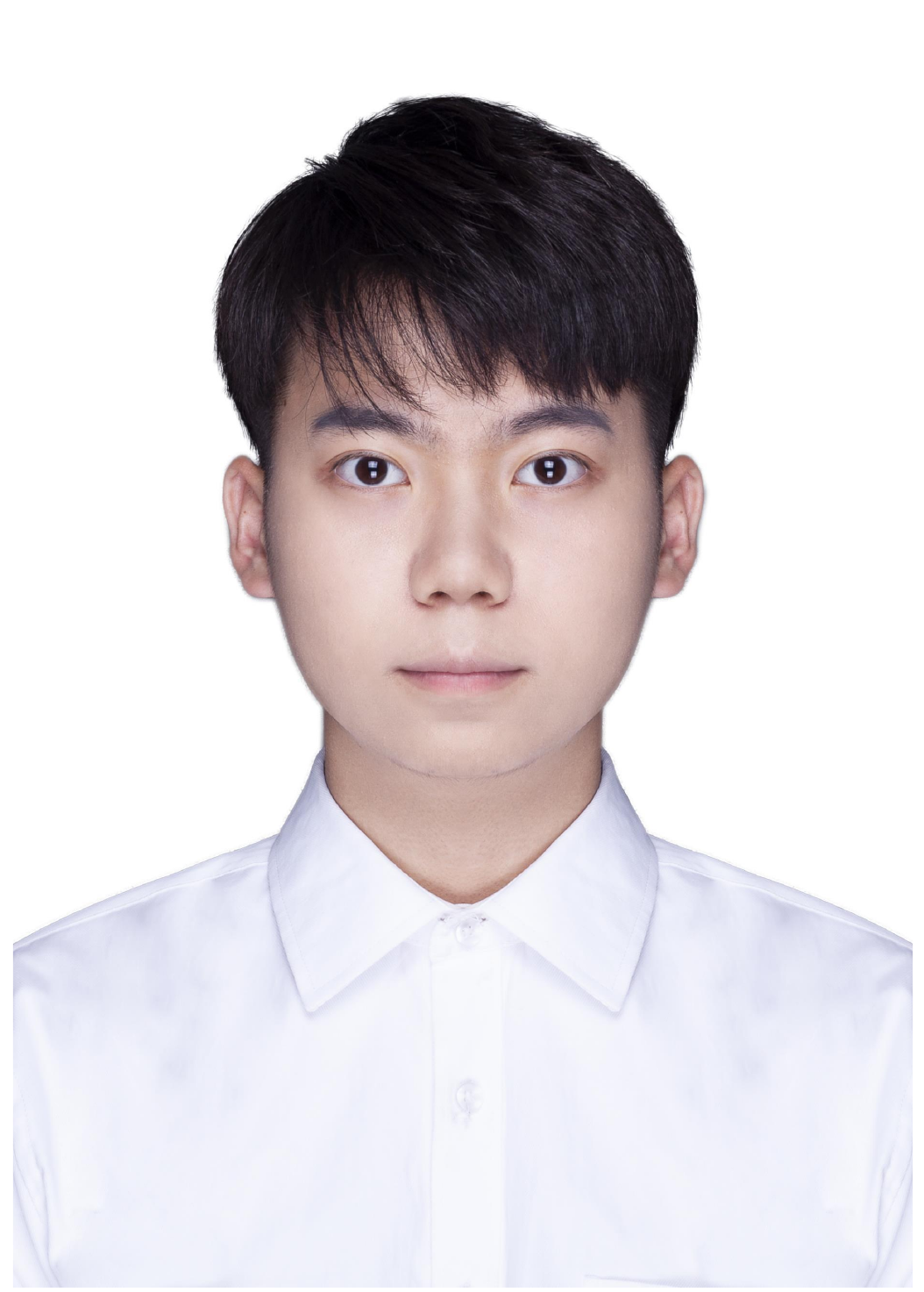}}]{Weixing Chen} has received the B.S. degree from the College of Medicine and Biological Information Engineering, Northeastern University, in 2020 and M.S. degree from Shenzhen Institute of Advanced Technology, Chinese Academy of Sciences in 2023. He is currently a Ph. D. student at the School of Computer Science and Engineering, Sun Yat-sen University. His main interests include embodied learning, multi-modal learning, and causal relation discovery. He has been serving as a reviewer for numerous academic journals and conferences, such as TNNLS, ECCV, ICCV, NuerIPS, ICLR, AAAI, MM.
\end{IEEEbiography}

\vspace{-16.5pt}
\begin{IEEEbiography}[{\includegraphics[width=1in,height=1.25in,clip,keepaspectratio]{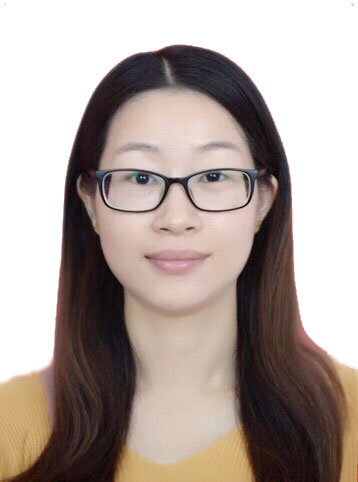}}]{Yongsen Zheng} received her Ph.D. degree in Computer Science and Technology from Sun Yat-sen University, Guangzhou, China, in 2023. She is currently a Research Fellow at the Nanyang Technological University, Singapore, and also at the National Centre for Research in Digital Trust, Singapore (Digital Trust Centre Singapore (DTC)), Singapore. Her current research interests include Human-AI Dialogue System, Conversational Recommender System, Natural Language Processing, Trustworthy AI, AI Safety, Knowledge Graphs, Large Language Models.
\end{IEEEbiography}

\vspace{-16.5pt}
\begin{IEEEbiography}[{\includegraphics[width=1in,height=1.25in,clip,keepaspectratio]{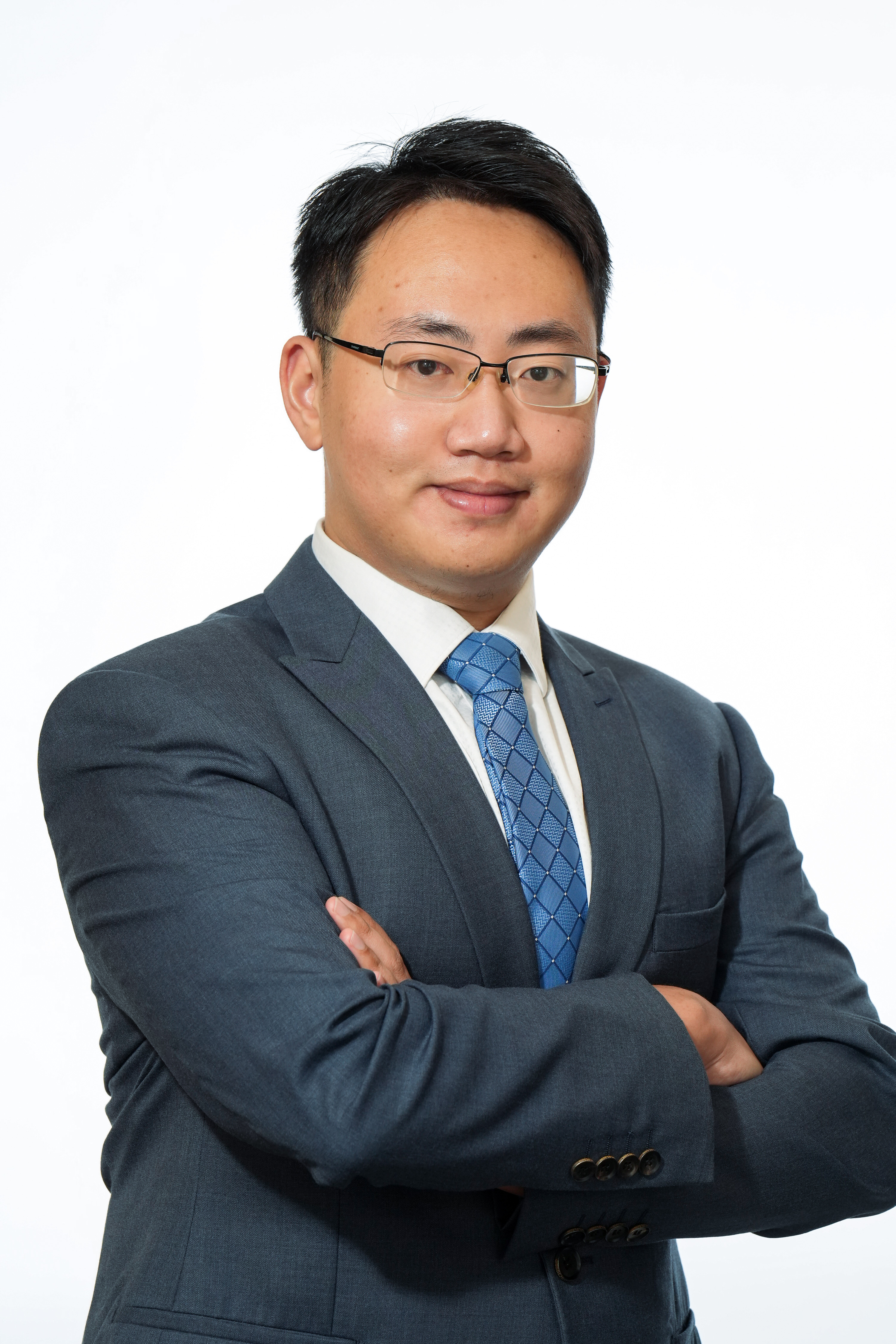}}]{Zechuan Chen} is now pursuing his Ph.D.degree at Sun Yat-Sen University. Before that, he was a research assistant at The University of Hong Kong (HKU), Hong Kong, China. He received the B.E. degree from Sun Yat-Sen University, China in 2022 and M.S. degree from The University of Hong Kong (HKU), Hong Kong, China in 2024. He was the winner of the 2022 International Student FinTech Innovation Competition. He is now leading FinTech-related research in the HCP Lab of Sun Yat-sen University. His research interests include FinTech, Multimodal-CoT, and ESG. He has served as an ICLR reviewer.
\end{IEEEbiography}

\vspace{-16.5pt}
\begin{IEEEbiography}[{\includegraphics[width=1in,height=1.25in,clip,keepaspectratio]{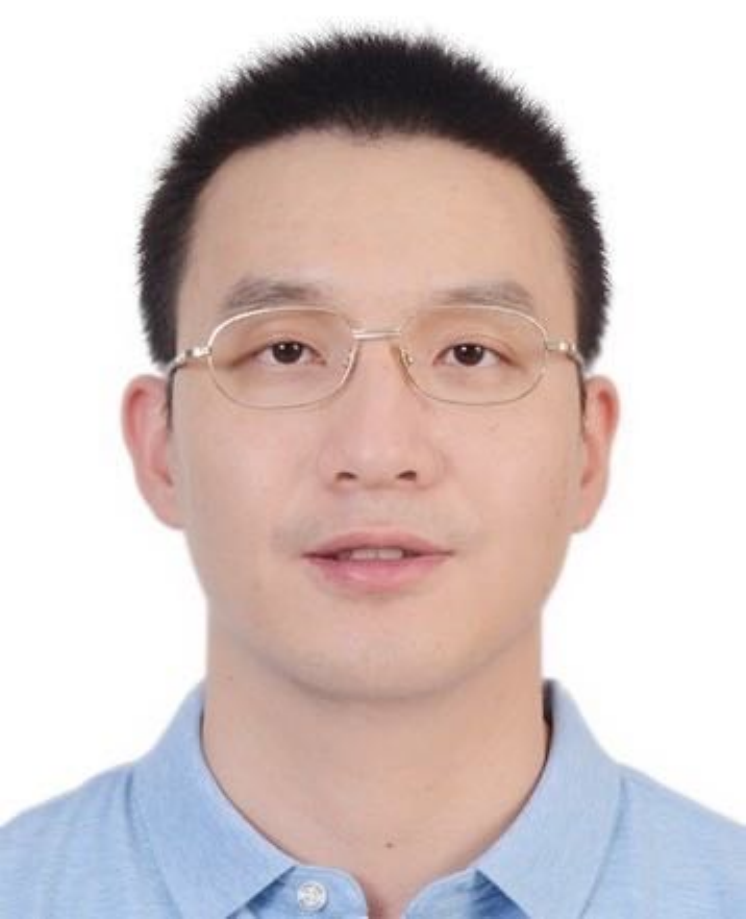}}]{Yang Liu}(M'21) is currently an associate professor at the School of Computer Science and Engineering, Sun Yat-sen University. He received his Ph.D. degree from Xidian University in 2019. His current research interests include multi-modal cognitive reasoning and causal relation discovery. He is the recipient of the First Prize of the Third Guangdong Province Young Computer Science Academic Show. He has authorized and co-authorized more than 30 papers in top-tier academic journals and conferences including TPAMI, TIP, TCSVT, CVPR, ICCV, IJCAI, and ACM MM. More information can be found on his personal website \hyperlink{https://yangliu9208.github.io.}{https://yangliu9208.github.io.}
\end{IEEEbiography}

\vspace{-16.5pt}
\begin{IEEEbiography}[{\includegraphics[width=1in,height=1.25in,clip,keepaspectratio]{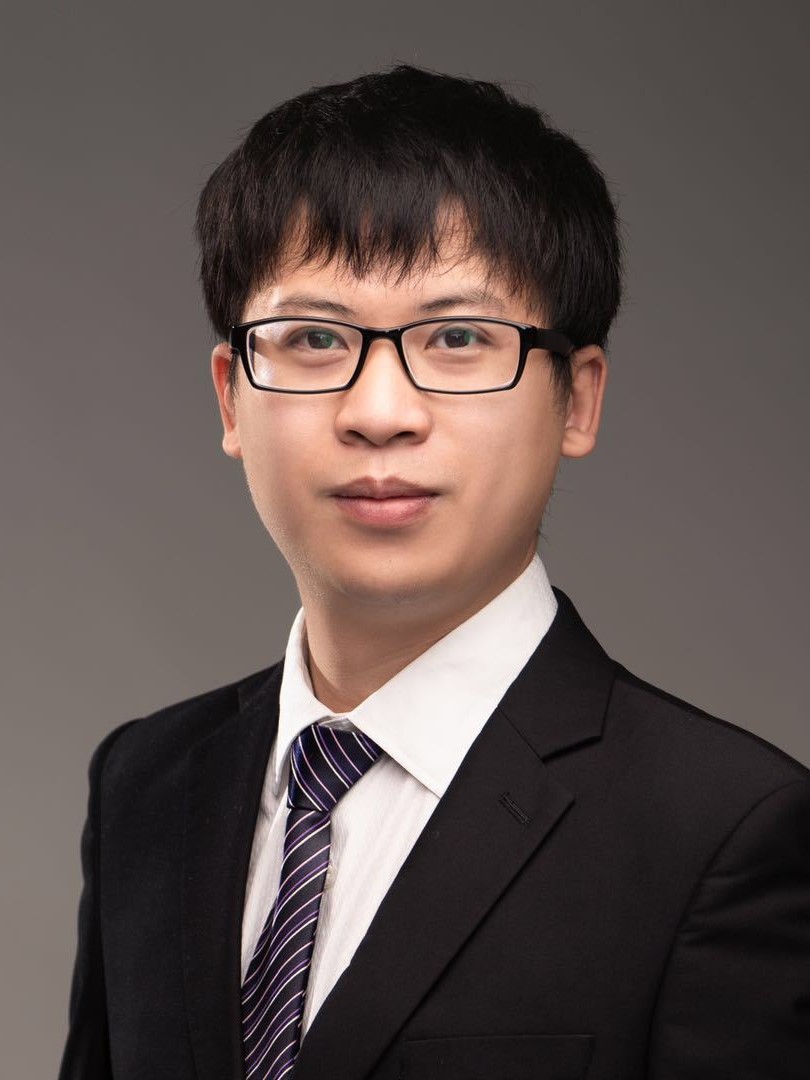}}]{Keze Wang} is nationally recognized as the Distinguished Young Scholars, currently serving as an Associate Professor at the School of Computer Science, Sun Yat-sen University, and a doctoral supervisor. He holds two Ph.D. degrees from Sun Yat-sen University (2017) and the Hong Kong Polytechnic University (2019). In 2018, he worked as a postdoctoral researcher at the University of California, Los Angeles, and returned to Sun Yat-sen University in 2021 as part of the "Hundred Talents Program." Dr. Wang has focused on reducing deep learning's dependence on training samples and mining valuable information from massive unlabeled data.
\end{IEEEbiography}

\vspace{-16.5pt}
\begin{IEEEbiography}[{\includegraphics[width=1in,height=1.25in,clip,keepaspectratio]{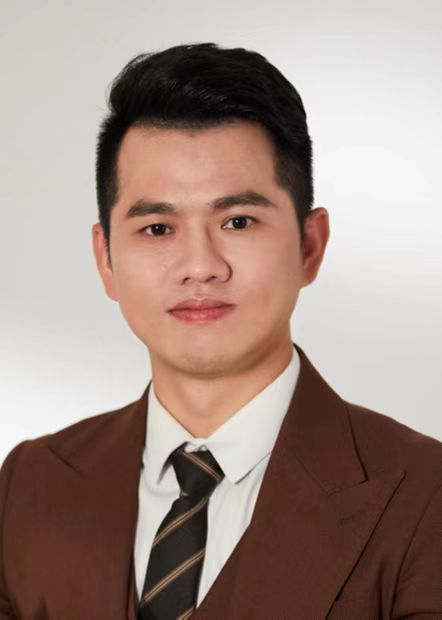}}]{Tianshui Chen} received a Ph.D. degree in computer science at the School of Data and Computer Science Sun Yat-sen University, Guangzhou, China, in 2018. Prior to earning his Ph.D, he received a B.E. degree from the School of Information and Science Technology in 2013. He is currently an associated professor in the Guangdong University of Technology. His current research interests include computer vision and machine learning. He has authored and coauthored more than 40 papers published in top-tier academic journals and conferences, including T-PAMI, T-NNLS, T-IP, T-MM, CVPR, ICCV, AAAI, IJCAI, ACM MM, etc. He has served as a reviewer for numerous academic journals and conferences. He was the recipient of the Best Paper Diamond Award at IEEE ICME 2017. 
\end{IEEEbiography}

\vspace{-16.5pt}
\begin{IEEEbiography}
[{\includegraphics[width=1in,height=1.25in,clip,keepaspectratio]{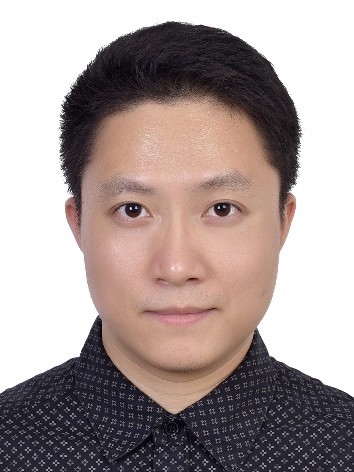}}]{Liang Lin}(Senior Member, IEEE) is a full professor of computer science with Sun Yat-sen University. He served as the executive director and distinguished scientist of SenseTime Group from 2016 to 2018, leading the R\&D teams for cutting-edge technology transferring. He has authored or co-authored more than 200 papers in leading academic journals and conferences, and his papers have been cited by more than 26\,000 times. He is an associate editor of \textit{IEEE Trans. Neural Networks and Learning Systems} and \textit{IEEE Trans. Multimedia}, and served as area chairs for numerous conferences, such as CVPR, ICCV, SIGKDD, and AAAI. He is the recipient of numerous awards and honors including Wu Wen-Jun Artificial Intelligence Award, the First Prize of China Society of Image and Graphics, ICCV Best Paper Nomination, in 2019, Annual Best Paper Award by Pattern Recognition (Elsevier), in 2018, Best Paper Dimond Award in IEEE ICME 2017, Google Faculty Award, in 2012. His supervised PhD students received ACM China Doctoral Dissertation Award, CCF Best Doctoral Dissertation and CAAI Best Doctoral Dissertation. He is a fellow of IET/IAPR.
\end{IEEEbiography}

\vfill

\end{document}